\documentclass{article}

\usepackage{microtype}
\usepackage{graphicx}
\usepackage{subfigure}
\usepackage{booktabs} 
\usepackage{bm}
\usepackage{enumerate}
\usepackage{algorithm}
\usepackage[noend]{algorithmic}
\usepackage{amsfonts}
\usepackage{xcolor}
\usepackage{bbm}
\usepackage{mathtools}
\usepackage{hyperref}

\usepackage[round]{natbib}

\usepackage[accepted]{icml2024}

\usepackage{amsmath}
\usepackage{amssymb}
\usepackage{mathtools}
\usepackage{amsthm}
\usepackage{footnote}
\usepackage[capitalize,noabbrev]{cleveref}

\theoremstyle{plain}
\newtheorem{theorem}{Theorem}[section]

\theoremstyle{definition}
\newtheorem{definition}[theorem]{Definition}

\theoremstyle{remark}

\usepackage[textsize=tiny]{todonotes}

\icmltitlerunning{A Bayesian Approach to Online Learning for Contextual Restless Bandits with Applications to Public Health}

\newcommand{\rev}[1]{{\color{black}#1}}
\newcommand{\revtwo}[1]{{\color{black}#1}}
\newcommand{\re}[1]{{\color{black}#1}}
\newcommand{\revaaai}[1]{{\color{black}#1}}
\begin{document}

\twocolumn[
\icmltitle{Context in Public Health for Underserved Communities: \\ A Bayesian Approach to Online Restless Bandits}




\begin{icmlauthorlist}
\icmlauthor{Biyonka Liang}{yyy}
\icmlauthor{Lily Xu}{harvardcs}
\icmlauthor{Aparna Taneja}{goog}
\icmlauthor{Milind Tambe}{harvardcs}
\icmlauthor{Lucas Janson}{yyy}
\end{icmlauthorlist}

\icmlaffiliation{yyy}{Department of Statistics, Harvard University, Cambridge, MA, USA}
\icmlaffiliation{goog}{Google Research India}
\icmlaffiliation{harvardcs}{Department of Computer Science, Harvard University, Cambridge, MA, USA}

\icmlcorrespondingauthor{Biyonka Liang}{biyonka@g.harvard.edu}

\icmlkeywords{Restless multi-armed bandits, online reinforcement learning, applied Bayesian modeling, hierarchical Bayes, Thompson sampling, public health, maternal health}

\vskip 0.3in
]



\printAffiliationsAndNotice{}  
\begin{abstract}
Public health programs often provide interventions to encourage program adherence, and effectively allocating interventions is vital for producing the greatest overall health outcomes, especially in underserved communities where resources are limited. Such resource allocation problems are often modeled as restless multi-armed bandits (RMABs) with unknown underlying transition dynamics, hence requiring online reinforcement learning (RL). We present Bayesian Learning for Contextual RMABs (BCoR), an online RL approach for RMABs that novelly combines techniques in Bayesian modeling with Thompson sampling to flexibly model the complex RMAB settings present in public health program adherence problems, namely context and non-stationarity.
BCoR's key strength is the ability to leverage shared information within and between arms to learn the unknown RMAB transition dynamics quickly in intervention-scarce settings with relatively short time horizons, which is common in public health applications. Empirically, BCoR achieves substantially higher finite-sample performance over a range of experimental settings, including a setting using real-world adherence data that was developed in collaboration with ARMMAN, an NGO in India which runs a large-scale maternal mHealth program, showcasing BCoR practical utility and potential for real-world deployment.
\end{abstract}

\section{Introduction}\label{sec:intro}

Public health programs in a wide array of areas such as communicable disease \citep{Killian_2019}, prenatal care \citep{armmanstudy1, ope2020reducing}, and cancer prevention \cite{wells2011community, lee2019optimal} often have many beneficiaries who are not adhering to their treatment. As adherence can be vital for ensuring positive health impacts, especially in the underserved communities these programs are designed to aid, programs frequently must decide how to allocate a scarce set of resources or \emph{interventions} to beneficiaries at risk of drop-out of the program due to continued non-adherence.
Such a constrained resource allocation problem is often modeled as a Restless Multi-Armed Bandit (RMAB) \citep{ogwhittle}, which is an extension of stochastic multi-armed bandits where each arm represents a Markov Decision Process (MDP). Specifically, a beneficiary is represented as an arm, their adherence status as the state of the corresponding MDP, and the allocation of an intervention as the action. The reward is a function of the adherrence status of the beneficiary (i.e., the state of the MDP) and a budget-constrained subset of arms are given intervention (i.e., pulled) at each timestep. Most relevantly for our work, RMABs have been used to model this resource allocation problem by our collaborators at ARMMAN \citep{armman}, an NGO in India which runs a large-scale 
maternal mHealth (mobile health) program \revaaai{that disseminates vital health information to pregnant beneficiaries via automated voice calls each week, with the goal of improving maternal and infant health outcomes \citep{mate, verma2023restless, ucw}. To encourage listenership (i.e., adherence), ARMMAN's healthcare workers can give live service calls (the intervention) to a subset of beneficiaries each week to troubleshoot potential barriers to adherence, thus improving overall listenership, which has been shown to result in to positive impact on the behavioural outcomes of the mothers \citep{dasgupta2024preliminarystudyimpactaibased}.} 

\revaaai{Developing an algorithm for resource allocation for mHealth programs like ARMMAN faces a few key challenges: (1)}~the transition dynamics of the underlying MDPs (e.g., corresponding to the beneficiaries' adherence) are unknown a priori; (2)~these settings are often resource scarce, so the (intervention) budget$~B$ is typically much smaller than the total number of arms$~N$; and (3)~the time horizon$~T$ is naturally limited to the treatment period, which is often small relative to~$N$. 
\re{For example, ARMMAN can only provide interventions to $\sim 2\%$ of their beneficiaries each week \cite{armmanstudy1}, and the time horizon is naturally limited by the duration of a pregnancy.} Due to the scarce intervention budget and relatively short time horizon, at any given time point, many (or most) of the beneficiaries in the program have never been intervened on. Thus, the algorithm must make a resource allocation decision despite not observing the underlying outcome distributions for a potentially vast number of arms.  
Additionally, previous analyses on public health programs indicate that beneficiary adherence varies with contextual factors \rev{such as income and education}, and that adherence rates can vary over time, suggesting non-stationarity in transition dynamics \citep{ mate, verma2023restless}. 
\re{However, these known properties of public health settings (e.g., short time horizon, contextual information, and non-stationarity) are largely unaddressed by existing online RL methods for RMABs. Neglecting these valuable properties which could improve an algorithm's ability to quickly learn an effective intervention allocation.}

\subsection{Main Contributions}
Driven by the needs of public health programs such as ARMMAN's maternal health program, we present \revtwo{\textbf{B}ayesian Learning for \textbf{Co}ntextual \textbf{R}MABs (BCoR), an online RL approach for RMABs which novelly combines techniques in hierarchical Bayesian modeling with Thompson sampling to account for the complexities of this application.}  
 In contrast with existing work on online RL for RMABs, which has largely focused on establishing asymptotic regret guarantees in $T$, we develop a learning approach that directly addresses an important real-world application, enabling known characteristics of the public health domain such as contextual information and non-stationarity to be incorporated for the first time. 
As such, a key component of our research approach is our direct collaboration with our stakeholders, ARMMAN, to understand the specific challenges around beneficiary adherence for public health program adherence. As adherence is crucial for many public health programs, which often share similar challenges as ARMMAN, our insights from ARMMAN enable us to precisely design and contextualize BCoR with respect to public health applications. 

Our research approach is aligned with other works which design bandit algorithms for real-world applications \citep{mary2015bandits, shen2015portfolio, ding2019interactive, bouneffouf2019survey} rather than providing improved theoretical results which, as we show in Section~\ref{sec:simulation_exp}, do not always correspond to improved practical performance. In fact, part of our contribution is pushing beyond the simplifying assumptions previous works have made to facilitate their theoretical analysis, enabling us to incorporate the complex structure present in applied examples; by necessity, this complexity makes theory more challenging and we consider it beyond the scope of this work.
In particular, BCoR's primary contribution is its strong empirical performance in experimental settings reflective of real-world settings:
BCoR achieves substantially higher reward than existing approaches across a wide array of realistic experimental settings, including a setting based on real data from ARMMAN's maternal health program and various settings where our model is misspecified. Such results exhibit BCoR's robustness and ability to leverage key characteristics of its application area, making it more suitable than existing approaches for real-world application. 

\section{Related Work}
\paragraph{Online RL for RMABs}
When the RMAB transition dynamics are \emph{known}, the Whittle index policy \citep{ogwhittle}, which pulls the top $B$ arms with the highest estimated future value if pulled (called the \emph{Whittle index}), asymptotically achieves the optimal time-averaged reward under certain conditions \citep{whittle_followup, wang2019opportunistic}. Since RMAB dynamics are \emph{unknown} in many applied settings, online RL approaches for RMABs generally use different learning techniques to quantify uncertainty on the transition probabilities\rev{, then apply a Whittle index policy, } 
for instance, by computing the Upper Confidence Bound (UCB) for each arm's state-action transitions \citep{ucw}, utilizing Thompson sampling \citep{jung2019regret, jung2019thompson, tswhittle} or Q-learning \cite{biswas2021, xiong2023finitetime}.
However, to our knowledge, all existing approaches for online RL in RMABs learn each arm's state-action transitions \textit{individually}, without sharing information within or between arms (e.g., via contextual information). Furthermore, not only are the asymptotic regret guarantees of these methods limited to simplified conditions as discussed above, but in addition, they usually only show 
empirical performance in relatively simple RMAB settings, such as when the number of arms~$N$ is small (usually $<100$) and the budget is high (usually $>30\%$ of $N$). 
ARMMAN and other real-world public health programs operate on a much larger scale, and we show in Section~\ref{sec:simulation_exp} that empirical performance in the aforementioned simple settings often does not translate to such realistic settings.

\paragraph{Incorporating Contextual Information}
Context is often present in bandit settings and can be highly informative \citep{hofmann2011contextual, boun2020survey}. 
While contextual information has been heavily explored in standard multi-armed bandits, e.g., \cite{auer2002using, langford2007contextbandit, chu2011contextual, li2020unifying}, there are no works, to our knowledge, which consider context for online learning in RMABs. 

\paragraph{Allowing for Non-Stationarity in RMABs}
While existing online RL methods for RMABs assume stationary transition dynamics \citep{biswas2021, gafni2022restless, ucw}, this assumption may not well approximate real-world settings \citep{mate, verma2023restless} and there is limited evidence to suggest that existing approaches are robust to non-stationarity \citep{biswas2021}. 
Though non-stationarity in RMABs has been explored for RMABs with known transition dynamics \citep{zayas2019asymptotically, indexnotenough, zhang2022near}, such solutions generally rely on solving a linear program directly using the true transition dynamics. It is unclear how such results could be extended to online RL settings where the algorithm must learn the transition dynamics and determine a good policy simultaneously.


\paragraph{Other Related Learning Approaches}
Learning RMAB transition dynamics can be considered a specific case of multi-task reinforcement learning (MTRL), which aims to learn the transition dynamics of a set of MDPs, often by modeling the MDPs as having some shared structure between them by, for instance, clustering MDPs with similar transition probabilities
\citep{wilson2007, lazaric2010, offlinedatasharing}. 
However, these MTRL approaches are largely designed for offline learning settings and, generally, do not consider contextual information and do not provide policy recommendations for regret minimization.  
\revaaai{For instance, past deployments of ARMMAN have largely focused on an offline learning approach precisely due to challenges this paper is set to address \citep{mate2020collapsing, mate}.}
Some online RL approaches focus on learning a single, sometimes partially observed, MDP 
However, these approaches are designed to learn a single MDP cannot be used to determine a set of actions to apply to a collection of MDPs, as in the RMAB setting. 




\section{Problem Setting}\label{sec:motivating_problem}
Consider an RMAB instance with $N$~arms. The learning algorithm interacts with the RMAB over $T$~timesteps with an (intervention) budget of $B \ll N$ pulls at each timestep, \rev{where$~T$ is fixed and known in advance.} Each arm is an MDP defined by the tuple $\left(\mathcal{S}, \mathcal{A}, R, \left\{P^{(t)}_{i}(s' \mid s, a)\right\}_{s', a, s, t}\right)$, where $\mathcal{S}$, $\mathcal{A}$, and $R: \mathcal{S} \times \mathcal{A} \to \mathbb{R}$ are the shared state space, action space, and reward function, respectively, across all arms and timesteps. The standard formulation for RMABs sets $\mathcal{A} = \{0, 1\}$ where $a=1$ represents a budget-constrained pull. The set of transition probabilities for arm~$i$ is $P_i \coloneqq \left\{P^{(t)}_{i}(s' \mid s, a)\right\}_{s', a, s, t}$, that is, for arm~$i$ in state $s \in \mathcal{S}$ that receives action $a \in \{0, 1\}$ at time $t \in [T]$, the transition probability to state $s' \in \mathcal{S}$ is $P^{(t)}_{i}(s' \mid s, a)$. Note that the transitions are indexed by time $t$, allowing for non-stationarity. We assume the reward function is known and the state of all arms is observable, even if they were not pulled. Importantly, we assume that all $P_i$ are \emph{unknown in advance} by the learning algorithm. Let $\bm{s}_{t} = (s_{1,t}, \ldots, s_{N,t})$ and $\bm{a}_{t} = (a_{1,t}, \ldots, a_{N,t})$ represent the tuple of states and actions across all arms at time~$t$, respectively, where we are constrained by $\sum_{i=1}^N a_{i, t} \leq B$ for all $t \in [T]$.
\revtwo{As in previous public health applications} \citep{ong2014effects, newman2018community, ayer2019prioritizing, lee2019optimal, mate, verma2023restless, ucw}, we also model program adherence with $\mathcal{S} = \{0, 1\}$, where state $s_{i,t} = 0$ represents beneficiary~$i$ being in a non-adhering state
, and $s_{i,t} = 1$ represents beneficiary~$i$ being in an adhering state. 
An action $a_{i,t} = 0$ represents no intervention on beneficiary $i$, and $a_{i,t}=1$ represents an intervention. A reward of $1$ is accrued when the beneficiary is in an adhering state and $0$ otherwise, i.e., $R(s, a) = s$.
Thus, the total reward at timestep~$t$ is a count of the number of beneficiaries who are in an adhering state, $\sum_{i=1}^N  s_{i,t}$,
and the time-averaged reward at timestep~$t$ (which we aim to maximize in this paper) is:
\begin{align}\label{eq:time_avg_reward}
    R^{(t)} = \frac{1}{t} \sum_{j=1}^t \sum_{i=1}^N  s_{i,j}.
\end{align}
Note that reward is calculated across all arms, as an arm may generate reward even when not pulled, i.e., a beneficiary may be in an adhering state even when no intervention is applied.
We assume the above state space, action space, and reward function in this paper, noting that the binary state and action spaces are largely for presentation purposes as it simplifies our notation and is standard in public health applications. BCoR's Bayesian modeling framework naturally extends to non-binary state and action spaces and general reward functions; see Appendix~\ref{sec:appendix_statespace}.

\section{The BCoR Algorithm}\label{sec:BCoR}

We introduce BCoR, which \revtwo{integrates a Bayesian model into Thompson sampling} for the online learning of RMABs with complex structure. \rev{We use \emph{hierarchical Bayesian modeling}, a Bayesian modeling approach where the prior distribution of some model parameters depends on other parameters, which are also assigned a priori. Hierarchical Bayesian models are flexible tools for modeling complex phenomena across broad application areas \citep{curry2013hierarchical, lawson2018bayesian, britten2021evaluating}, as the hierarchical structure on the model parameters can be used to represent complex relationships and interactions between variables of the model.

}
\subsection{Learning the Transition Dynamics}\label{sec:learning}
To apply Thompson sampling, we must specify a Bayesian model of the RMAB's reward distribution. Since our rewards equal our states, we focus on modeling the state transition distribution, \rev{i.e., the $P^{(t)}_i(1 \mid s, a)$'s, for all $s \in \{0, 1\}, a\in \{0, 1\}$, $t \in [T]$, and $i \in [N]$.\footnote{Note, $P^{(t)}_i(0 \mid s, a)=1-P^{(t)}_i(1 \mid s, a)$ deterministically, so we only need to model the $P^{(t)}_i(1 \mid s, a)$ transitions.}} To specify this model, we will first consider the simple non-contextual RMAB with stationary transitions and incrementally add complexity, separately explaining each addition until our full model is presented.
\re{The flexibility of hierarchical Bayesian modeling often comes with a corresponding computational cost, and hence the components we discuss in this section are carefully chosen not just to incorporate properties of our applied setting, but also to maintain computational tractability; see Appendix~\ref{sec:appendix_sim} for further details.}

\paragraph{Sharing Information Within an Arm}
A \rev{simple and }natural \rev{ choice of Bayesian model} for this simple RMAB is {to treat }
$P_i(1 \mid s, a)$ as drawn \rev{independently} from some distribution \rev{(e.g., $\text{Unif}[0, 1]$)}, where we remove the superscript for time (for now). 
Hence, this model aims to learn each arm's state-action transitions \emph{individually} --- requiring the model to learn $4N$ different transition probabilities (i.e., each of the four state-action pairs for each arm). \rev{This learning approach may not be very effective} because, as discussed in Section~\ref{sec:intro}, the budget and time horizon may be small relative to $N$, and hence, there may be many arms for which the algorithm never observes behavior under $a=1$. However, since the vast majority of arms will receive $a=0$ at each timestep, we expect to observe a relatively large set of outcomes for each arm when $a=0$ over time. Through discussions with ARMMAN, we also expect that, for a given arm $j$, its transition dynamics when $a=0$ have some relationship to its transition dynamics when $a=1$. Thus, it can be useful to \textit{share information within an arm} to better estimate an arm's active ($a=1$) transition probabilities, for which there is very little data, based on its passive ($a=0$) transition data, for which there is much more data. Hence, we propose to model this relationship as:
\begin{align}
\begin{split}
  P_i(1 \mid s, 0)& = \Phi\left(\alpha_i^{(s, 0)}\right) \\ \label{eq:re_model_share}
   P_i(1 \mid s, 1) &= \Phi\left(\alpha_i^{(s, 1)} + b_0\alpha_i^{(0, 0)} + b_1\alpha_i^{(1, 0)} \right)
\end{split}
\end{align}
for all $s \in \mathcal{S}$, where $\Phi$ is the standard normal cumulative distribution function. \rev{Here, $b_0$, $b_1$, and each of the ${\alpha_i^{(s,a)}}$'s are parameters of this Bayesian model, and, as is standard in Bayesian models of this form \cite{gelman2013bayesian}, we will set their priors as zero-centered Normal distributions.} Hence, we can interpret the ${\alpha_i^{(s,0)}}$'s as representing \rev{each arm's} individual passive transitions ($a=0$), the $\alpha_i^{(s,1)}$'s as representing the active transitions ($a=1$), and $b_0$ and $b_1$ as representing  \rev{the informativeness of }
\rev{an arm's transition dynamics under passive actions for inferring its dynamics under active action}. Hence, the parameters $b_0$ and $b_1$ enable us to use information about passive actions, of which we observe many, to inform our inference on active actions, of which we observe very few.
As we {set the prior on $b_0$ and $b_1$ to be zero-centered}, which corresponds to no information sharing, our model will only learn to share information if \revtwo{the data suggests that} such a relationship exists. 


\paragraph{Sharing Information Across Arms}
The ideas presented above deal with sharing information \textit{within} a given arm. As RMAB problems often come with contextual information for each arm, \revtwo{such as age, education, and other demographic factors}, it is desirable to use this information to share information \textit{across} arms. For instance, we may reasonably expect that arms with similar covariates will have similar behavior \revtwo{(e.g., low-income beneficiaries tend to have lower adherence \citep{mohan2021can}).} Given a covariate matrix $\bm{X} \in \mathbb{R}^{N \times k}$ where the row vectors $X_i$ represent feature vectors for each of the arms, we incorporate $\bm{X}$ into Model~\eqref{eq:re_model_share} by adding a parameter $\bm{\beta}^{(s, a)} \in \mathbb{R}^k$ for each state-action pair and modeling the transitions as:
\begin{align}
  P_i(1 \mid s, 0)& = \Phi\left(X_i\bm{\beta}^{(s, 0)} + \alpha_i^{(s, 0)}\right) \nonumber\\
   P_i(1 \mid s,1) &= \Phi\Big(X_i\bm{\beta}^{(s, 1)} + \alpha_i^{(s, 1)} \\
   & \qquad \qquad \qquad + b_0 \alpha_i^{(0, 0)} + b_1 \alpha_i^{(1, 0)}\Big), \nonumber
\end{align}
where, similar to $b_0$ and $b_1$, we set \rev{zero-}centered Normal priors on the $\bm{\beta}^{(s, a)}$'s.
Note, the four $\bm{\beta}^{(s, a)}$ vectors are shared \textit{across} all arms for \textit{each} state-action pair. \rev{Since we may not observe many transitions when $a=1$ due to budget constraints, it will be harder to learn the $\bm{\beta}^{(s, a=1)}$'s.} To facilitate learning the $\bm{\beta}^{(s, a=1)}$'s, we can \rev{\emph{add a level of hierarchy} to the $\bm{\beta}^{(s, a)}$'s by modeling} all four of the $\bm{\beta}^{(s, a)}$ vectors as having the same mean vector $\bm{\mu}_{\bm{\beta}}$, \rev{which is a new parameter in our model, on which we} place a (normally distributed) prior; see the third and sixth lines of Model~\eqref{final_model} for the explicit forms of $\bm{\mu}_{\bm{\beta}}$ and the $\bm{\beta}^{(s, a)}$'s. 
Intuitively, our posterior updates of $\bm{\mu}_{\bm{\beta}}$ would use data across all arms\rev{' state-action transitions} to learn where the covariate effects $\bm{\beta}^{(s, a)}$'s are ``centered'' and our posterior updates of each $\bm{\beta}^{(s, a)}$ would use data across arms specifically in state~$s$ that receive action~$a$ to learn 
\rev{how far that } particular $(s, a)$ pair \rev{deviates from the center}. \rev{Hence, \emph{this hierarchy facilitates greater information sharing}.} 

\paragraph{Addressing Non-Stationarity}
As is standard in the RMAB literature, we have so far treated the transition dynamics as stationary or fixed over time. However, in real-world scenarios, the arms often exhibit non-stationary transition dynamics \citep{mate, verma2023restless}. To model these time effects, we use spline regression, a common approach for flexibly modeling non-linear effects \citep{esl}. Given a spline basis matrix $\bm{M} \in \mathbb{R}^{T \times d}$ with rows $\rev{M_t}$, where $T$ is the time horizon and $d$ is the dimension of the spline basis, we can incorporate non-stationarity into our model as:
    \begin{align}
  P_i^{(t)}(1 \mid s, 0)& = \Phi\left(X_i\bm{\beta}^{(s, 0)} + \rev{M_t}\bm{\eta}^{(s, 0)} + \alpha_i^{(s, 0)}\right)\nonumber \\
   P_i^{(t)}(1 \mid s, 1) &= \Phi\Big(X_i\bm{\beta}^{(s, 1)} + \rev{M_t}\bm{\eta}^{(s, 1)} +  \alpha_i^{(s, 1)} \\ & \qquad \qquad \qquad \qquad + b_0\alpha_i^{(0, 0)} + b_1\alpha_i^{(1, 0)}\Big), \nonumber
\end{align}
where we set \rev{zero-}centered Normal priors on the $\bm{\eta}^{(s, a)}$'s and  we now have superscripts on the $P_i^{(t)}(s'\mid s, a)$ to denote time-varying transition dynamics. \rev{Hence, $\bm{\eta}^{(s, a)}$ represents the magnitude of the time effects on the transition dynamics.}

Lastly, \rev{we place a prior on the variance of the $\alpha_i^{(s, a)}$'s, which we denote $\tau^2_{\alpha^{(s, a)}}$; see the fourth and fifth lines of Model~\eqref{final_model} for the explicit forms of the $\tau^2_{\alpha^{(s, a)}}$'s and $\alpha_i^{(s, a)}$'s.} We do so because, without this prior, the $\alpha_i^{(s, a)}$'s would be modeled \textit{per arm}, while all other parameters 
are shared across arms. Hence, we cannot directly use the posteriors of the $\alpha_i^{(s, a)}$'s to infer anything about new arms \rev{since they only represent information about a single arm's state-action pair}. Adding a prior \rev{on the variance} enables us to share information across arms for all parameters. 
See Definition~\eqref{final_model} for the full statement of the BCoR learning model. We let $\bm{0}_k \in \mathbb{R}^k$ be the $k$-dimensional vector with all $0$ entries and $I_{k \times k}$ be the $k \times k$ identity matrix. 

\rev{Hence, the user-specified values which are required as inputs to our model are: $\tau_0, \sigma_0, \tau_{\bm{\mu}}^2$, $\tau_{b_0}^2$, $\tau_{b_1}^2$, $\tau^2_{\bm{\beta}^{(s, a)}}$ and $\tau^2_{\bm{\eta}^{(s, a)}}$, for all $s \in \{0, 1\}, a \in \{0, 1\}$.} \rev{Such user inputs are common in Bayesian modeling, and, as more data is observed, 
the posterior distributions of the parameters will be most strongly influenced by the actual data rather than these specific inputs \citep{vdv, gelman2013bayesian}. \revtwo{The input values used in all experimental results in this paper were chosen to ensure they were reasonably default values given the problem setting; see Appendix~\ref{sec:appendix_sim} for the exact specification and further discussion. Importantly, we used the \emph{same} input values for \emph{all} experimental results presented in this paper, including our example constructed from real ARMMAN data and many misspecified simulation settings where these input values do not correctly reflect the RMAB's true structure,  across various configurations of$~N$,$~T$ and$~B$. Our experimental results across these various settings, shown in Figures~\ref{sim_exp}--\ref{armman_exp} and Figures~\ref{sim_exp_all}--\ref{armman_exp_all}, suggest that BCoR is primarily learning from the data and hence, the specific input values had little impact on the performance of the algorithm.}}

Hence, using insights from our collaboration with domain experts, we carefully incorporate characteristics of our application area into the structure of BCoR using hierarchical Bayesian modeling, which has not previously been used for online learning in the RMAB literature. 
\re{Though it may seem limiting to assume a linear model, Bayesian linear models 
have an extensive history of being sufficiently expressive and empirically effective across a wide range of complex settings that almost surely do not satisfy linearity \citep{gelman2007analysis, hilbe2009data, curry2013hierarchical, gelman2013bayesian, lawson2018bayesian, britten2021evaluating}.
From a bias-variance tradeoff perspective, a linear model is more suitable for the small sample size settings we consider compared to to a more complex model, which would have higher variance. Notably, the real data-based example of Section~\ref{sec:real_data_exp} is highly misspecified and strongly violates the linearity assumptions, and BCoR is still achieves higher reward than existing approaches.
}

\subsection{Online Arm Selection}
To apply Thompson sampling, we can update the posterior distribution of our model parameters at each timestep, and take a draw from the posterior. 
As we observe more data over time, we expect the posterior distributions of our model parameters to concentrate around values \rev{that best fit the data}, and hence, so will our estimates of the transition dynamics. 

\begin{definition}[The BCoR Learning Model]\label{final_model}
    \begin{align}
b_0 &\sim \mathcal{N}\left(0, \tau_{b_0}^2\right) \nonumber\\
b_1 &\sim \mathcal{N}\left(0, \tau_{b_1}^2\right) \nonumber\\
\bm{\mu}_{\bm{\beta}} &\sim   \mathcal{N}\left(\bm{0}_k, \tau_{\bm{\mu}}^2I_{k \times k}\right) \nonumber \\
\tau^2_{\alpha^{(s, a)}} & \sim \mathrm{Inv}\text{-}\mathrm{Gamma}(\tau_0, \sigma_0) \qquad \rev{\forall s, a}  \nonumber\\
 \alpha_i^{(s, a)} &\sim \mathcal{N}\left(0, \tau^2_{\alpha^{(s, a)}}\right) \qquad  \rev{\forall s, a}\nonumber\\
  \bm{\beta}^{(s, a)} &\sim \mathcal{N}\left(\bm{\mu}_{\bm{\beta}}, \tau^2_{\bm{\beta}^{(s, a)}}I_{k \times k}\right) \qquad  \rev{\forall s, a} \\
    \bm{\eta}^{(s, a)} &\sim \mathcal{N}\left(\bm{0}_d, \tau^2_{\bm{\eta}^{(s, a)}}I_{d \times d}\right) \qquad  \rev{\forall s, a} \nonumber\\
  P_i^{(t)}(1 \mid s, 0)& = \Phi\left(X_i\bm{\beta}^{(s, 0)} + \rev{M_t}\bm{\eta}^{(s, 0)} + \alpha_i^{(s, 0)}\right)\nonumber \\
   P_i^{(t)}(1 \mid s, 1) &= \Phi\Big(X_i\bm{\beta}^{(s, 1)} + \rev{M_t}\bm{\eta}^{(s, 1)} +  \alpha_i^{(s, 1)}   \\
   & \qquad \qquad \qquad \qquad +b_0\alpha_i^{(0, 0)} + b_1\alpha_i^{(1, 0)}\Big), \nonumber
\end{align}
\end{definition}

 Concretely, for all $s  \in \{0, 1\}$, $a \in \{0, 1\}$, $i\in [N]$, let $\tilde{b}^{(t)}_0, \tilde{b}^{(t)}_1$, $\tilde{\alpha}_i^{(s, a)(t)}$, $\tilde{\bm{\eta}}^{(s, a)(t)}$, $\tilde{\bm{\beta}}^{(s, a)(t)}$ represent a draw from the posterior distributions of $b_0, b_1$, $\alpha_i^{(s, a)}$, $\bm{\eta}^{(s, a)}$, $\bm{\beta}^{(s, a)}$ at time $t$, respectively.
We can generate estimates of the transition probabilities $\tilde{P}^{(t)}_i\left(1 \mid s, a \right) $ by plugging these posterior draws into the last two lines of Model~\eqref{final_model}. 
Specifically, for all $s  \in \{0, 1\}$, $a \in \{0, 1\}$, $i\in [N]$,
   \begin{align}\label{eq:p_i}
   \tilde{P}^{(t)}_i(1 | s, 0) &\coloneqq \Phi\Big(X_i\tilde{\bm{\beta}}^{(s, 0)(t)}
    + M_t\tilde{\bm{\eta}}^{(s, 0)(t)} + \tilde{\alpha}_i^{(s, 0)(t)}\Big) \\
   \tilde{P}_i^{(t)}(1 \mid s, 1) &\coloneqq \Phi\Big(X_i\tilde{\bm{\beta}}^{(s, 1)(t)} + M_t\tilde{\bm{\eta}}^{(s, 1)(t)} +\tilde{\alpha}_i^{(s, 1)(t)}\\
   & \qquad \qquad \qquad + \tilde{b}^{(t)}_0\tilde{\alpha}_i^{(0, 0)(t)} + \tilde{b}^{(t)}_1\tilde{\alpha}_i^{(1, 0)(t)}\Big). \nonumber
   \end{align}
Using the $\tilde{P}^{(t)}_i\left(1 \mid s, a \right)$'s, we implement 
a Whittle index policy \rev{\citep{ogwhittle}}, which computes the Whittle index using the set of all $\tilde{P}^{(t)}_i\left(1 \mid s, a \right)$'s and pulls the $B$ arms with the highest Whittle indices. 
See Definition~\ref{def:whittle-index} in Appendix~\ref{sec:appendix_sim}
\begin{savenotes}
\begin{algorithm}[htb]
\caption{BCoR}
\label{alg:BCoR}
\begin{algorithmic}[1]
\STATE \textbf{Input:} $N$~arms, budget~$B$, time horizon~$T$, covariate matrix~$\bm{X} \in \mathbb{R}^{N \times k}$, spline basis matrix~$\bm{\rev{M}} \in \mathbb{R}^{T \times d}$, model inputs \rev{\{$\tau_0, \sigma_0, \tau_{\bm{\mu}}^2$, $\tau_{b_0}^2$, $\tau_{b_1}^2$, $\tau^2_{\bm{\beta}^{(s, a)}}$, $\tau^2_{\bm{\eta}^{(s, a)}}\}$ for all $s \in \{0, 1\}, a \in \{0, 1\}$.}
\FOR{timestep $t$ $ \in \{1, \ldots, T\}$}
\STATE 
Observe $\bm{s}_t$ and use all historical data to compute the posterior distribution of Model~\eqref{final_model}'s parameters.\footnote{At time $t=1$, before having observed transitions, the posterior remains the prior.}
\STATE \rev{From the posterior distribution computed in the previous step,} generate $\tilde{P}^{(t)}_i\left(1 \mid s, a \right)$ \revtwo{as in Equation~\eqref{eq:p_i}} for all $s \in \{0, 1\}, a \in \{0, 1\}, i \in [N]$. 
\STATE \rev{Using the $\tilde{P}^{(t)}_i(1 | s, a)$'s generated in the previous step}, compute the Whittle index for all $i \in [N]$ 
\rev{and pull the $B$ arms with the highest Whittle indices}.
\ENDFOR
\end{algorithmic}
\end{algorithm}
\end{savenotes}
 for a formal definition of the Whittle index \citep{ogwhittle}.
The Whittle index is computable via an efficient binary search approach presented in \cite{qian2016restless}. Further details on the implementation and computational efficiency of BCoR are in Appendix~\ref{sec:appendix_sim}.
\section{Experiments}\label{sec:experiments}
We show that BCoR consistently achieves high reward across various experimental settings, 
even in challenging settings where the data generating model is misspecified.
We also evaluate performance in a setting constructed from a real-world public health campaign, namely ARMMAN's maternal healthcare program. General implementation details for all experiments in this paper are in Appendix~\ref{sec:appendix_sim}, with details specific to Sections~\ref{sec:simulation_exp} and~\ref{sec:real_data_exp} in Appendices~\ref{sec:appendix_sim_only} and~\ref{sec:appendix_armman}, respectively. \revtwo{The code, data, and instructions needed to reproduce the results \rev{of Section~\ref{sec:simulation_exp}, as well as all additional simulations in Appendix~\ref{sec:appendix_sim_results}}, are available via our Github: \texttt{https://github.com/biyonka/BCoR}.}


\subsection{Methods Under Comparison}
We evaluate the BCoR algorithm as described in Algorithm~\ref{alg:BCoR}. 
 For comparison, we consider the \textit{UCWhittle} approach of \citet{ucw} (denoted \textit{UCW-Value} in their paper). This approach, which computes a UCB for each arm's state-action transitions and selects an ``optimistic'' value within the confidence bound to plug into the Whittle index policy, exhibits superior empirical performance over other existing approaches such as \citet{biswas2021} and \citet{wang2019opportunistic}. 
We also consider a Thompson sampling--based approach based on \citet{tswhittle}, denoted \textit{TS}, which performs Thompson sampling on the \emph{individual} arm's state-action pairs (i.e., it models each arm's state-action transitions individually with no information sharing), then plugs the estimated transitions into the Whittle index policy. 
For baselines, the \textit{Random} algorithm assigns $a=1$ to $B$ arms uniformly at random 
at each timestep, 
providing a lower baseline for the reward without the use of any learning approach. 
We also implement \rev{a} Whittle index \emph{oracle} approach, which executes the Whittle index policy using the true transition dynamics. 
\begin{figure*}[h]
  \centering
\includegraphics[width=\linewidth]{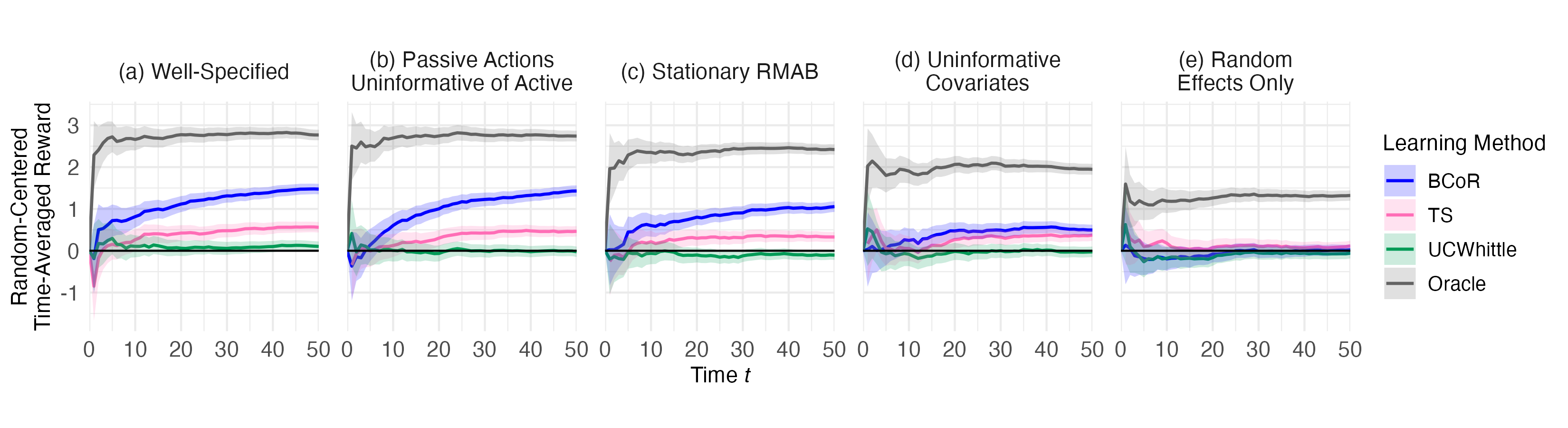}
\vspace{-2em}
\caption{
\rev{We generate all RMAB instances using $N=400,T=50$, and$~B=10$, i.e.,$~B$ is $2.5\%$ of$~N$, across $1{,}000$ random seeds.} The covariate matrix $\bm{X}$ is randomly generated with $k=4$ (two continuous covariates and two categorical) across the random seeds. \rev{The various RMAB} simulation settings \rev{are detailed in Section~\ref{sec:simulation_exp} and can be summarized as} (a)~a well-specified setting \rev{(no components of  Model~\eqref{final_model} are zero'ed out)}, (b)~a setting where passive actions are uninformative of active actions ($b_0=b_1=0$), (c)~a stationary setting ($\bm{\eta}^{(s,a)} =\rev{\bm{0}, \forall s,a}$), (d)~a setting with uninformative covariate information ($\bm{\mu}_{\bm{\beta}}=0$, $\bm{\beta}^{(s,a)} =0, \rev{\forall s,a}$), and (e)~a \rev{highly misspecified} setting\rev{, i.e., one where the RMAB instances are stationary with no information sharing between or within the arms}. Lines represent the time-averaged reward of each method \rev{averaged over the $1{,}000$ random seeds} with the Random baseline subtracted out. Error bars depict $\pm 2$ SEs. 
}
  \label{sim_exp}
\end{figure*}

\begin{figure*}[h]
  \centering
\includegraphics[width=\linewidth]{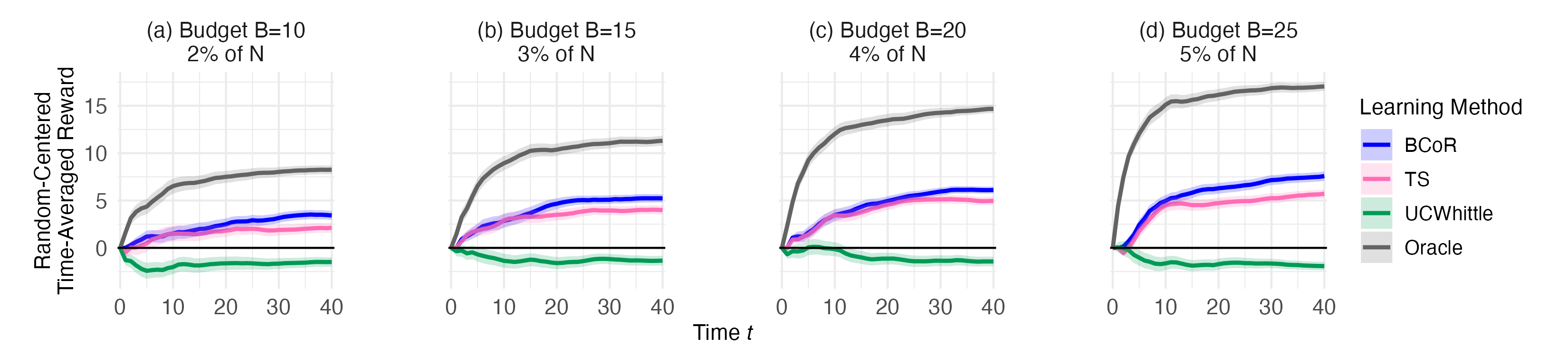}
\caption{
Performance of various methods on the ARMMAN data-driven example described in Section~\ref{sec:real_data_exp} with $N=500, T=40$, and varying budget$~B$, where all $B\leq 5\%$ of $N$ to reflect the magnitude of real-world budget constraints.  Lines represent the time-averaged reward of each method averaged over $100$ random seeds with the Random baseline subtracted out. \rev{Note, the grey line is an oracle approach with access to the true transitions.} Error bars depict $\pm 2$ SEs. \rev{UCWhittle performs worse than random across all settings, which can sometimes occurs when the budget is relatively small and the time horizon is short, \revtwo{though it recovers over a longer time horizon}; see Figure~\ref{fig:ucwhittle_worse}.} 
}
  \label{armman_exp}
\end{figure*}


\subsection{Simulation Experiments}\label{sec:simulation_exp}
Given a fixed number of arms~$N$, time horizon~$T$, and budget~$B$, we use Model~\eqref{final_model} to generate simulated RMAB instances over $1{,}000$ random seeds. \rev{For each instance, we run all algorithms and calculate} 
the time-averaged reward (Equation~\ref{eq:time_avg_reward}) at each timestep $t \in [T]$. The initial state provided for each algorithm is randomized across the seeds. \rev{We plot the average performance across the $1{,}000$ seeds, as shown in Figure~\ref{sim_exp}.}

We explore various parameterizations of Model~\eqref{final_model} to generate a well-specified setting and various misspecified settings for the BCoR learning model. For the well-specified setting of Figure~\ref{sim_exp}(a), the parameterization of Model~\eqref{final_model} used to generate the RMAB instances is the same as the prior used for BCoR. \rev{The misspecified settings shown in Figures~\ref{sim_exp}(b--d) each represent zeroing out just one component of Model~\eqref{final_model} to generate the RMAB instances (and in each setting; all components not explicitly zero'ed out are left as in Model~\eqref{final_model}).} 
\rev{For instance, Figure~\ref{sim_exp}(b) and (c) represent a setting }where the transitions are truly stationary (i.e., we set $\bm{\eta}^{(s,a)} =0, \forall s,a$ \rev{when generating the RMAB instances across the random seeds}), but the prior for BCoR \rev{never changes from the one used in the well-specified setting}. Hence, the prior allows for properties like non-stationarity and informative contextual information, and BCoR must learn \revtwo{from the data} that some of these properties are not present.
\rev{In particular, Figure~\ref{sim_exp}(e) represents a} highly misspecified setting where the transition probabilities are generated \rev{by zeroing out all components of Model~\eqref{final_model} and just leaving the random effects $\alpha_i^{(s, a)} \sim \mathcal{N}(0, \sigma^2)$ (note we also remove the \revtwo{prior on the variance of the} $\alpha_i^{(s, a)}$'s), so that the transition dynamics are just generated as $ P_i(1 \mid s, a)=\Phi\left(\alpha_i^{(s, a)}\right)$ for all $s, a$. }
In such a setting, the transitions are \emph{stationary} and there is \emph{no information sharing} within an arm or across arms. While existing approaches such as UCWhittle and TS are implicitly designed for this setting (since they learn each arm's state-action transitions individually), BCoR must learn that the RMAB instances have no information sharing and are stationarity. Hence, this setting is particularly challenging for BCoR. See Appendix~\ref{sec:appendix_sim} and~\ref{sec:appendix_sim_only} for further details about the simulation environment.

In the well-specified setting of Figure~\ref{sim_exp}(a) and the partially misspecified setting of Figures~\ref{sim_exp}(b--c), BCoR achieves significantly higher reward than other approaches. \revaaai{For instance, in Figure~\ref{sim_exp}(c), BCoR's final time-averaged reward is \emph{more than double} that of the next-best solution, TS, with even larger engagement wins in the settings of  Figures~\ref{sim_exp}(a)--(b), potentially correspond to significant positive impacts in overall health outcomes in real-world deployment.} 
In particular, Figures~\ref{sim_exp}(a--c) have covariate structure, showing that even when the RMAB is stationary, which is the setting TS and UCWhittle are designed for, ignoring informative covariate information when present can significantly decrease performance. 
In Figure~\ref{sim_exp}(d), BCoR achieves slightly higher reward than the other approaches by accounting for the non-stationarity, even though it has the additional challenge of learning that the covariates are completely uninformative. 
In Figure~\ref{sim_exp}(e), it is only possible to learn each arm's state-action transitions individually. 
In this setting, none of the methods perform significantly better than random over the entire time horizon. These results show that if no structure is present,  and the time horizon~$T$ and budget~$B$ are small relative to $N$, the learning problem is too challenging for essentially any approach to significantly outperform random. Intuitively, this is because the only way to learn about a particular state-action transition is to directly observe it, but having small $T$ and $B$ relative to $N$ means that any algorithm will only observe a small portion of the RMAB's dynamics. Hence, in applied settings such as ARMMAN's maternal health program, where we expect similar configurations of $T$, $B$, and $N$, it is essential to use a learning algorithm that can leverage properties present in the RMAB instance.
\emph{We repeated this experiment for different RMAB configurations, varying the number of arms$~N$, the time horizon$~T$, the  budget$~B$, and the number of covariates$~k$, 
which are in Appendix~\ref{sec:appendix_sim_results}. Those results show similar trends as in Figure~\ref{sim_exp}, exhibiting BCoR's robustness to these different experimental settings. } \revtwo{Additionally, recall from Section~\ref{sec:learning} that we used the \emph{same} prior for \emph{all} experimental results in this paper, where each plot represents an average over $1{,}000$ different RMAB instances. BCoR's performance in misspecified settings across these many RMAB instances suggests that it is not very sensitive to the specific prior used and is effectively learning from the data.\footnote{For instance, if BCoR was highly sensitive to its prior, it would not be able to effectively learn from the data when the RMAB is, e.g., stationary. However, BCoR achieves high reward in this setting (see Figure~\ref{sim_exp}(b)) suggesting BCoR is not highly sensitive.}} 
In summary, these experiments exhibit how existing approaches, 
\re{which have ostensibly strong theoretical guarantees, can perform close to, and sometimes no better than, a random selection algorithm in moderate sample regimes, even in simplified settings amenable to their theoretical guarantees such as when stationarity is present. These empirical results exhibit that the types of existing theoretical guarantees found in restless bandit papers are insufficient to ensure high performance in public health applications.} 
While all methods will improve as they observe more samples (over $T$), only BCoR can use information over the $N$ arms, thus allowing it to learn more quickly and efficiently in challenging settings where $T$ and $B$ are limited.




\subsection{\revtwo{Experiment Using Real Data From ARMMAN}}\label{sec:real_data_exp}

\revaaai{It is essential to evaluate the performance of BCoR experimentally on a data-driven simulator before running it on actual ARMMAN beneficiaries in order to confirm its expected performance before actual deployment in real-world settings. Hence,} we construct a data-driven simulator that approximates the true dynamics based on \rev{real} historical ARMMAN covariate data, leveraging our extensive collaborations with ARMMAN to inform the design of the simulator. ARMMAN provided anonymized covariate information from $24{,}011$ beneficiaries enrolled in their maternal health program, collected in 2022. We generate the data\rev{-driven} simulator by using ARMMAN's internal estimates of the true transition probabilities given a beneficiary's covariate information. We choose $N$, $T$, and $B$ to reflect the learning challenges present in the ARMMAN setting, e.g., $T=40$ because that is the approximate length of a pregnancy in weeks, and the varying budget values are reflective of ARMMAN's true budget constraints.

 As shown in Figure~\ref{armman_exp}, BCoR achieves the highest reward across all budget constraints, translating to significant increases in overall engagement compared to the next best performing method; \revaaai{for instance, a \emph{$61\%$ increase in engagement} in the tightest, \revaaai{and hence most realistic,} budget setting ($B=10$).}
\revaaai{Hence, BCoR enables RMABs to be applied to realistic public health settings with significantly better performance at unprecedented scale in $N$ and shorter horizon $T$, potentially leading to life-saving health outcomes in real-world mHealth settings.} See Appendix~\ref{sec:appendix_armman} for further details on implementation and data privacy protocols.

\section{Conclusion and Ethical Considerations}\label{sec:conclusion}
 We present BCoR, the first online RL approach for contextual and non-stationary RMABs \revaaai{designed for mHealth in close collaboration with domain experts at ARMMAN. }
Using a novel combination of techniques in Bayesian hierarchical modeling combined with Thompson sampling, BCoR outperforms existing approaches \revaaai{across a wide range of challenging empirical settings reflective of real-world applications, 
significantly increasing the potential social impact of RMABs for real-world resource allocation.}
\revaaai{Importantly, BCoR is designed to improve 
listenership in mHealth settings 
and hence, \emph{would not} withhold any health information from beneficiaries. Our data-driven simulator is considered secondary analysis, was approved by ARMMAN's ethics board, and was performed with fully anonymized data collected with consent prior to data collection. See Appendix~\ref{sec:broader_impact} and~\ref{sec:appendix_armman} respectively for discussion on the ethical considerations of BCoR.
}

\section*{Acknowledgements}
L.X. was supported by a Google PhD fellowship and B.L. was partially supported by the NSF Graduate Research Fellowship Program. L.J. was partially supported by the NSF grant CBET-2112085. The authors would like to thank Jun Liu for helpful discussion on our Bayesian model and Nathan Cheng and anonymous reviewers for their thoughtful comments. 

\bibliography{example_paper}
\bibliographystyle{icml2024}

\appendix
\section*{Technical Appendix}

\section{Broader Impact}\label{sec:broader_impact}
BCoR provides an online learning approach to intervention allocation in public health programs for improved beneficiary adherence. 
Adherence in public health programs has been shown to improve target measures of health in areas such as sexual/reproductive health \citep{elazan2016reproductive}, cardiac disease management \citep{corotto2013heart}, cancer prevention \citep{wells2011community}, and infant and maternal health in the case of ARMMAN \cite{armmanstudy1, mohan2021can, kilkaristudy1, mate}. In particular, these programs often operate in low-resource communities where healthcare workers may be scarce or lack the resources to provide individualized care to a large proportion of their beneficiaries. Approaches like BCoR can be used to improve intervention allocation for these programs to generate larger overall health improvements for their target populations.

One advantage of BCoR over black-box models such as \cite{mate2020collapsing} is that the user can generate diagnostic measures from the posterior distribution to assess how strongly the individual covariates and potential non-stationarity are impacting the model's predictions. These diagnostic measures enable administrators to interpret the model outputs (e.g., seeing which covariates most strongly influence the final prediction), thus allowing them to screen for potentially unfair intervention allocations. Such results could be used to inform improvements, and even new iterations, of BCoR, thus enabling more fair and responsible deployment. 
\rev{Additionally, since BCoR shares information across all arms for all parameters, it can immediately use the posterior distribution of the parameters based on previously observed arms to infer the transition dynamics of new arms. This can aid in resource allocation in certain enrollment settings where new beneficiaries join partway through ongoing programs; see Appendix~\ref{sec:appendix_cont_enroll} for additional discussion.}

To contextualize the potential real-world impact of our approach, we provide some additional details about ARMMAN. ARMMAN is a non-profit based in India that runs mMitra, a mobile health program that disseminates vital health information to pregnant beneficiaries via automated voice calls each week, with the goal of improving maternal and infant health outcomes. To encourage listenership, ARMMAN's community healthcare workers can give live service calls (the intervention) to a subset of beneficiaries each week. These live calls allow the community healthcare workers to troubleshoot potential barriers to information access, thus improving the beneficiaries' listenership (i.e., adherence) to the program.

\revtwo{In the context of deployment for ARMMAN, the BCoR algorithm would be used to enhance the allocation of live service calls to beneficiaries, which is \emph{in addition} to the automated calls all ARMMAN beneficiaries already receive. Hence, in practice, our algorithm \emph{would not withhold any health information} from beneficiaries. As described in \cite{ucw}, all beneficiaries receive the same weekly health information via ARMMAN's automated calls, regardless of who receives a live service call, and no additional vital health information is provided in live service calls that is not already provided in the automated service calls. Hence, the same health information is available to all beneficiaries regardless of who receives a live service call. In addition to these scheduled live service calls, beneficiaries can request service calls themselves via a free missed call service provided by ARMMAN. If deployed, our algorithm would only help with the \emph{scheduled} live call allocation and would not hinder access to these requested service calls. Importantly, note, we \textit{did not actually} deploy BCoR or any other method to ARMMAN's actual beneficiaries in the analyses of this paper.}


\section{Extensions to More General State Spaces, Action Spaces, and Reward Functions}\label{sec:appendix_statespace}
The focus on binary state and action spaces is largely for presentation purposes as it simplifies our exposition and notation, and is standard in public health applications \citep{ong2014effects,  ayer2019prioritizing, mate, verma2023restless, ucw}. In fact, BCoR's Bayesian learning framework naturally extends to larger (discrete) state and action spaces. For instance, to include an additional action (denoted $a=2$), we can easily incorporate the expanded action space into BCoR by, for instance, modeling:

 $$ P_i^{(t)}(1 \mid s, a=2) = \Phi\Big(X_i\bm{\beta}^{(s, 2)} + M_t\bm{\eta}^{(s, 2)} +  \alpha_i^{(s, 2)}\Big),$$

for all states $s$, and adding other terms relating $ P_i^{(t)}(1 \mid s, a=2)$ to the other state-action pairs as appropriate based on domain knowledge of the $a=2$ action (similarly to what we did for $a=1$). The same extension could be applied to expanding the state space. Implementing such changes would essentially just require us to incorporate the single line above into the Bayesian model fitting code via the \texttt{Rstan} package in \texttt{R}. Thus, though BCoR as implemented exactly in our code may not be totally general for larger state and action spaces, our framework is general, and the modifications for accommodating such generality only involve changing a few lines of code. 

\subsection{General Implementation Details}\label{sec:appendix_sim}


The following implementation details apply to all experimental results presented in this paper, including those in Sections~\ref{sec:simulation_exp} and~\ref{sec:real_data_exp}, and all related experimental results in the following Appendix sections.

\re{\subsubsection{Computing the Whittle Index Policy}
We first provide the formal definition of the Whittle Index \citep{ogwhittle}, as adapted from \cite{ucw}:
\begin{definition}[Whittle index]  \label{def:whittle-index}
Given transition probabilities~$P_i$, a state~$s$, and a discount factor $\gamma \in (0, 1)$, the \emph{Whittle index}~$W_i(P_i, s)$ of arm~$i$ in state $s$ is defined as $  W_i(P_i, s) \coloneqq \inf\limits_{m_i} \{ m_i : Q^{m_i}(s, 0) = Q^{m_i}(s, 1) \}$
where the Q-function $Q^{m_i}(s,a)$ and value function $V^{m_i}(s)$ are the solutions to the Bellman equation with penalty~$m_i$ for pulling action $a=1$: 
\begin{align}
\begin{split}
    Q^{m_i}(s, a) &= -{m_i} a + R(s, a) + \gamma \sum\limits_{s' \in \mathcal{S}} P_i(s' \mid s, a) V^{m_i}(s')\label{eqn:whittle-index}\\
    V^{m_i}(s) &= \max\limits_{a \in \mathcal{A}} Q^{m_i}(s, a) \ . 
    \end{split}
\end{align}
\end{definition}

To compute the Whittle Index policy, we use the binary search-based implementation provided in the code repository for \cite{ucw}, which incorporates a number of computational speed-ups for computing the Whittle Index policy, such as a memoizer to avoid making repeated calculations and an early termination scheme that stops the Whittle index computation for an arm whenever an upper bound on its Whittle index is less than the top $B$ Whittle indices already computed (since the only way an arm could be pulled is if it has a Whittle index higher than the current top $B$ arms).
As mentioned previously, the key metrics of interest to ARMMAN assess how our algorithm will perform on their data, for which our best proxy without having actually deployed our algorithm on real people, is the types of simulations we present in the paper. In our simulations, we find that the Whittle index computation of \cite{ucw} is fast; for instance, running UCWhittle for the $N=400, T=50$ settings of Figure 1 took $\approx 0.7-0.8$ seconds using a single CPU, on average. Hence, as we elaborated on further in the following response, our algorithm, which uses this computationally efficient implementation of the Whittle index policy, is able to meet the run-time demands of its intended real-world application.
}

\re{\subsubsection{Updating the BCoR Posterior}
As is standard in Bayesian modeling \citep{neal2011mcmc, betancourt2017conceptual, matamoros2020introduction, stan}, we generate our posterior updates using Hamiltonian Monte Carlo, a Markov chain Monte Carlo (MCMC) approach of generating a sequence of samples that converges to some target distribution (i.e., our posterior) when direct sampling is difficult. Quantifying the time complexity of MCMC methods like HMC remains an open problem, as the number of Monte Carlo evaluations needed for posterior sampling to converge is not known a priori \citep{neal2011mcmc, betancourt2017conceptual, matamoros2020introduction}. However, our implementation of Bayesian model fitting in Stan incorporates a number of best practices to improve computational speed, as described in \citep{stan}.

Importantly, a key characteristic of public health settings is that the resource allocation policy does not need to be updated continuously e.g., for ARMMAN, the allocation of live calls is updated each week \citep{mobilehealth1}, and so, only a single update of the posterior is needed a single time each week. As mentioned previously, the key metrics of interest to ARMMAN assess how our algorithm will perform on their data, for which our best proxy without having actually deployed our algorithm on real people, is the types of simulations we present in the paper. In our simulations, we find that the computation time needed for BCoR to make a posterior draw was $10$ minutes at maximum, e.g.,  in the settings shown in Figure~\ref{sim_exp}, where $N=400$, BCoR took $\approx 9-10$ minutes to generate a posterior sample at the end of the time horizon $T=50$ using a single CPU, which is is the most computationally intensive posterior update to compute since it incorporates all $N*(T-1)$ samples previously observed. While some existing approaches, which for instance rely on very simple Bayesian models with closed-form posteriors that are easy to sample from (e.g., basic Beta-Binomial models), may be computationally faster, BCoR is still advantageous over such approaches in terms of the tradeoff between performance and computational efficiency for our setting of interest. Waiting a few minutes or even hours for a resource allocation assignment that only needs to happen once a week is computationally efficient relative to the total time available for computation and hence, would not be logistically prohibitive to implement in practice. In discussions with ARMMAN representatives, they emphasized that having an effective resource allocation is of higher priority than having the fastest possible algorithm. Hence, our algorithm is able to provide a highly effective resource allocation (e.g., achieving a $>60$\% increase in reward compared to the next best approach as shown in Figure~\ref{armman_exp}) while still meeting the run-time demands of its intended real-world application.

}

\subsubsection{The BCoR Prior}
\begin{figure*}[h]
  \centering
  \subfigure[]{\includegraphics[width=0.4\linewidth]{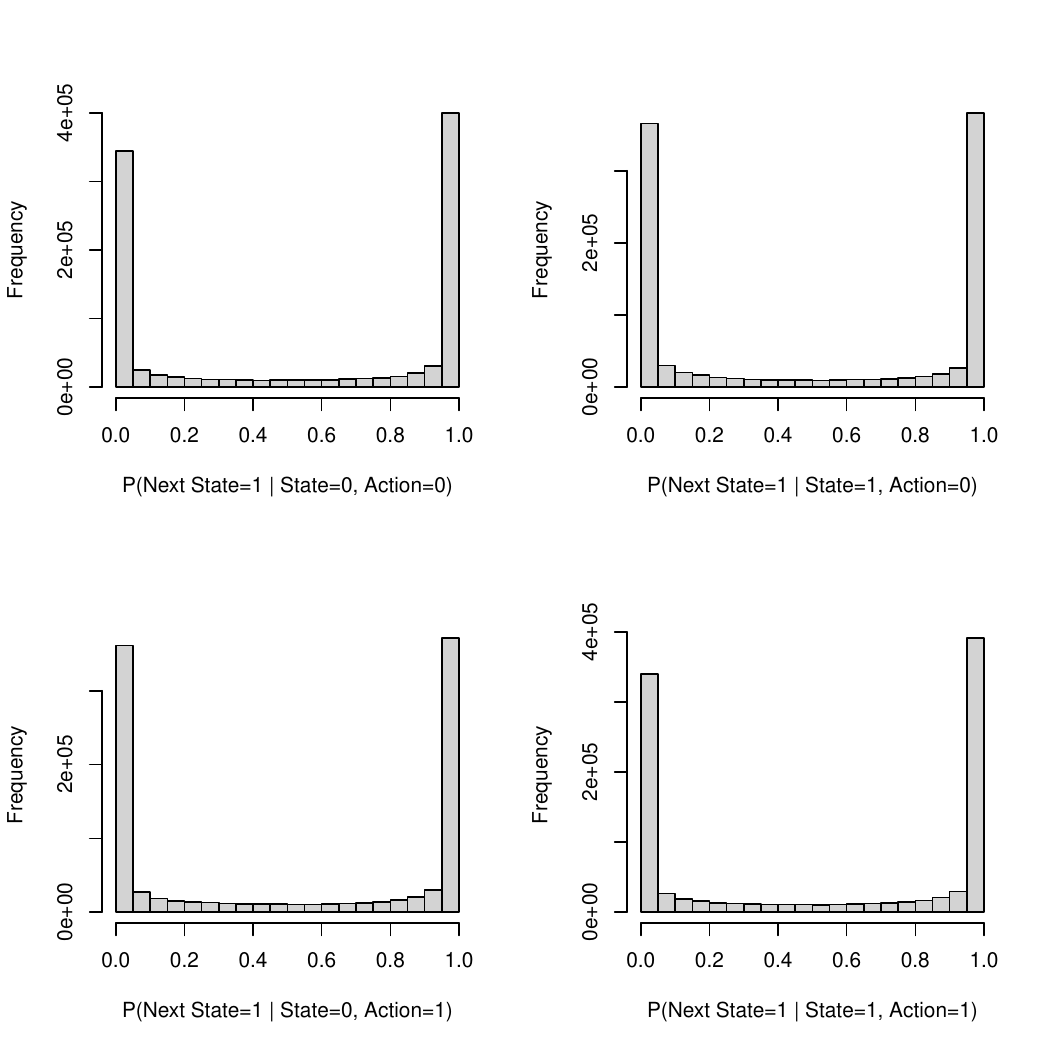}}
  \subfigure[]{\includegraphics[width=0.4\linewidth]{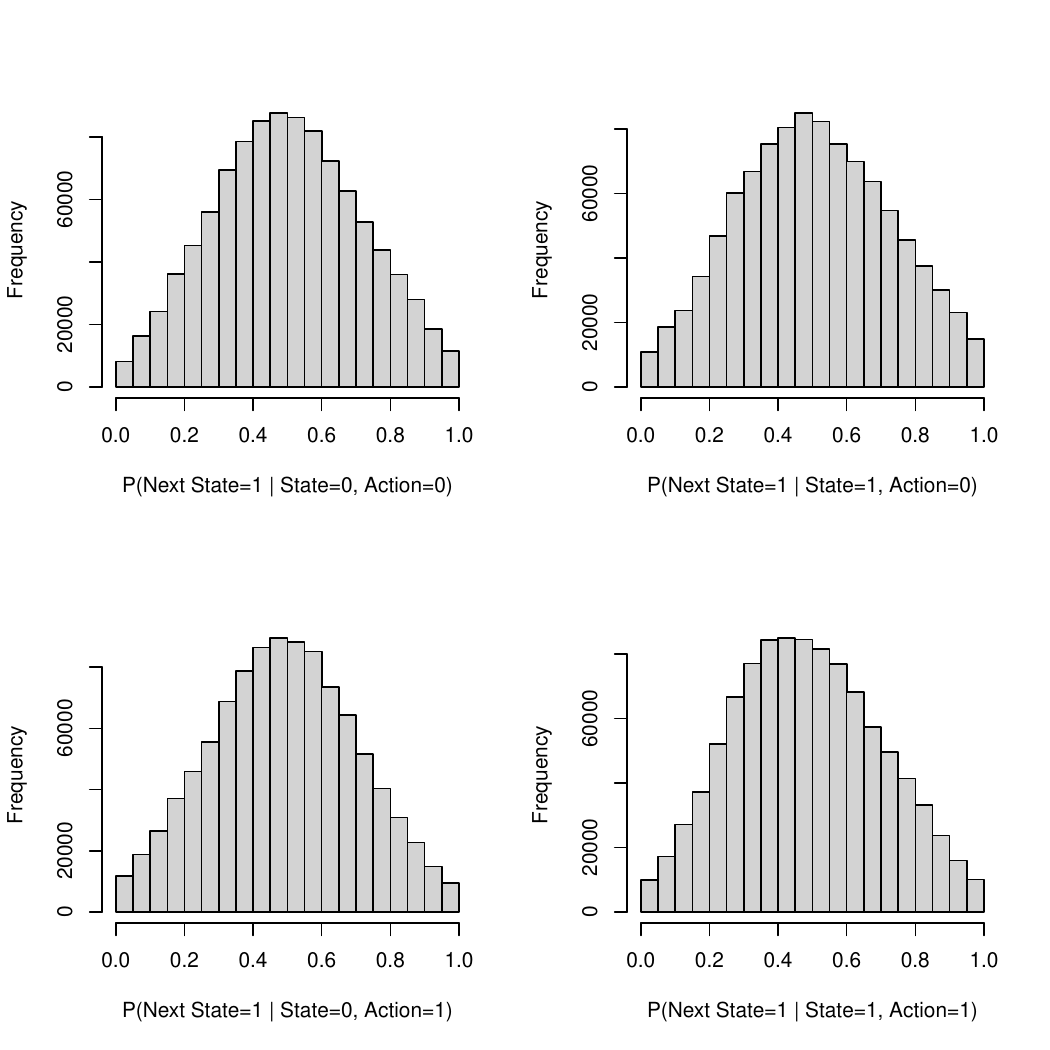}}
\caption{Implied priors on transition probabilities using (a) a wide prior on the model parameters,  
$b_0 \sim \mathcal{N}\left(0, 2^2 \right),
b_1 \sim \mathcal{N}\left(0, 2^2 \right),
\bm{\mu}_{\bm{\beta}} \sim   \mathcal{N}\left(\bm{0}_k, 2^2 I_{k \times k}\right),
\tau^2_{\alpha^{(s, a)}} \sim \text{Inv-Gamma}(100, 1),
  \bm{\beta}^{(s, a)} \sim \mathcal{N}\left(\bm{\mu}_{\bm{\beta}}, 2^2 I_{k \times k}\right), $ and $\bm{\eta}^{(s, a)} \sim \mathcal{N}\left(\bm{0}_d, 2^2 I_{d \times d}\right)$, and (b) the prior specified in Model~\eqref{prior}. Hence, in (b), the prior variances are set much wider than what was used for the experimental results in this paper (represented by (a)). Histograms show transitions probabilities for RMAB instances with $N=400, T=50, B=10$ generated using Model~\eqref{final_model} across $50$ random seeds. The covariate matrix $\bm{X}$ and the spline matrix $\bm{B}$ are generated as described in Section~\ref{sec:appendix_sim_only}.  Note that when using a wide prior, the transition probabilities tend to concentrate around $0$ and $1$, which is not representative of most realistic examples. }
  \label{fig:priors}
\end{figure*}

For all experimental results in this paper, we use the following prior specification for the BCoR model:
    \begin{align}\label{prior}
b_0 &\sim \mathcal{N}\left(0, 0.1^2 \right) \nonumber\\
b_1 &\sim \mathcal{N}\left(0, 0.1^2 \right) \nonumber\\
\bm{\mu}_{\bm{\beta}} &\sim   \mathcal{N}\left(\bm{0}_k, 0.3^2 I_{k \times k}\right) \nonumber \\
\tau^2_{\alpha^{(s, a)}} & \sim \text{Inv-Gamma}(100, 1)  \nonumber\\
 \alpha_i^{(s, a)} &\sim \mathcal{N}\left(0, \tau^2_{\alpha^{(s, a)}}\right)\\
  \bm{\beta}^{(s, a)} &\sim \mathcal{N}\left(\bm{\mu}_{\bm{\beta}}, 0.1^2 I_{k \times k}\right) \nonumber\\
    \bm{\eta}^{(s, a)} &\sim \mathcal{N}\left(\bm{0}_d, 0.3^2 I_{d \times d}\right) \nonumber
\end{align}
It may seem more natural to set wide priors on the parameter to encompass a larger possible range of RMAB instances, but setting wide priors on the parameters generate implied transition dynamics which concentrate around $0$ and $1$, as shown in Figure~\ref{fig:priors}a.

From discussions with ARMMAN representatives, and as reflected in the transitions generated from our real data simulator in Figure~\ref{fig:real_data_transitions}, it is standard to assume that most beneficiaries to have some probability of switching to an engaging state that is somewhere in the middle of the $[0, 1]$ range, e.g., $[0.1, 0.9]$, with a smaller proportion of beneficiaries with transitions closer to $0$ and $1$. Hence, we chose the prior specification of Model~\eqref{prior} because it generates implied priors on the transition probabilities that reflect this default expected behavior, see Figure~\ref{fig:priors}b. This prior also provides a more fair comparison with our other Bayesian approach, \textit{TS}, which we initialize with a uniform prior on $[0, 1]$ to take advantage of Beta-Binomial conjugacy for the posterior updates. The BCoR learning model (Model~\ref{final_model}) was fit using the \texttt{rstan} package \citep{stan} in the computing language \texttt{R}.

\subsubsection{The Whittle Oracles}
We now provide additional discussion on the Whittle index oracle used in our experimental results. Note, though the Whittle oracle approach presented in Section~\ref{sec:experiments} has access to the full true RMAB transition probabilities, non-stationarity cannot be easily incorporated in the Bellman equation 
(Equation~\eqref{eqn:whittle-index} of Definition~\ref{def:whittle-index}) for computing the Whittle index. Hence, to implement a Whittle-index-based oracle, we must summarize the potential non-stationarity in different ways.
We implement three versions of Whittle index oracles. The \textit{Current Time Whittle Oracle} computes the Whittle index using each arm's state-action transitions at the current time point. The \textit{Time Average Whittle Oracle} uses the average of the transition dynamics for a given arm's state-action pair across all time. Finally, recall that the \textit{Cumulative Average Whittle Oracle} uses the average of the transition dynamics \textit{up to the current time} for a given arm's state-action pair. Intuitively, all three Whittle oracles perform a version of the Whittle index policy, but handle the (possible) non-stationarity in different ways. Note, in a stationary setting, all three Whittle oracles reduce to the same method, which is the standard Whittle index policy for stationary RMABs. We found that all three oracles performed comparably in all simulation settings as well as our real data setting. Hence, for clarity of presentation, we only show one of the Whittle oracles, the \textit{Cumulative Average Whittle Oracle}, in the results of Figures~\ref{sim_exp} and~\ref{armman_exp} of Sections~\ref{sec:simulation_exp} and~\ref{sec:real_data_exp}.  

For all methods under comparison, including the Whittle oracles, we compute Whittle indices using a discount of $\gamma = 0.9$ as in \cite{ucw}. 

\subsubsection{A Greedy Policy}
 Additionally, we implement versions of the TS and BCoR methods with a Greedy policy instead of a Whittle policy, where the greedy policy calculate the estimated treatment effect of action on arm $i$ at time $t$ as: 
$$\text{TE}^{(t)}_i\left(s_{i,t}\right) \coloneqq \tilde{P}^{(t)}_i\left(1 \mid s_{i,t}, 1\right) -\tilde{P}^{(t)}_i\left(1 \mid s_{i,t}, 0\right),$$
and pull the top $B$ arms with the highest $\text{TE}^{(t)}_i(s_{i,t})$.
We also implement the analogous \textit{Greedy Oracle}, which pulls the top $B$ arms with the highest $\text{TE}^{(t)}_i(s_{i,t})$ using the true transition probabilities.

\subsubsection{Additional Details}
For each each random seed, we randomly initialize the starting state vector for all methods under comparison. \rev{The initial action vector for BCoR is randomly initialized, i.e., at $t=1$, before any data has been observed, we randomly select $B$ arms to assign $a = 1$.} For implementing UCWhittle in our experiments, we used the implementation provided in the code repository of \cite{ucw}.

\subsection{Additional Details on Section~\ref{sec:simulation_exp}}\label{sec:appendix_sim_only}
In the simulations of Section~\ref{sec:simulation_exp}, the covariate matrix $\bm{X} \in \mathbb{R}^{400 \times 4}$ is generated with $k=4$ simulated covariates, so that for each arm $i =1, ..., 400$, we had:
\begin{align*}
    X_{i,1} &\sim \text{round}(\mathcal{N}(22, 2^2))\\
    X_{i,2} & \sim \mathcal{N}(0, 1^2)\\
    X_{i,3} & \sim \mathcal{N}(0, 1^2)\\
    X_{i,4} & \sim \text{Bern}(0.5)
\end{align*}
The $X_{i,1}$'s represent a simulated age, the $X_{i,2}$'s and $X_{i,3}$'s represents some mean-centered and normalized continuous covariate, and the $X_{i,4}$'s represent a binary categorical covariate. The $X_{i,1}$'s were mean centered and standardized before being used for data generation. A P-spline matrix of degree three, implemented using the \texttt{ps} function of the \texttt{dlnm} package in the computing language \texttt{R} was used to generate non-stationarity \citep{dlnm}. Knots were automatically selected as described and recommended in the documentation of the \texttt{ps} function in \texttt{dlnm}.
 
\begin{figure*}[htb!]
  \centering
\includegraphics[width=\linewidth]{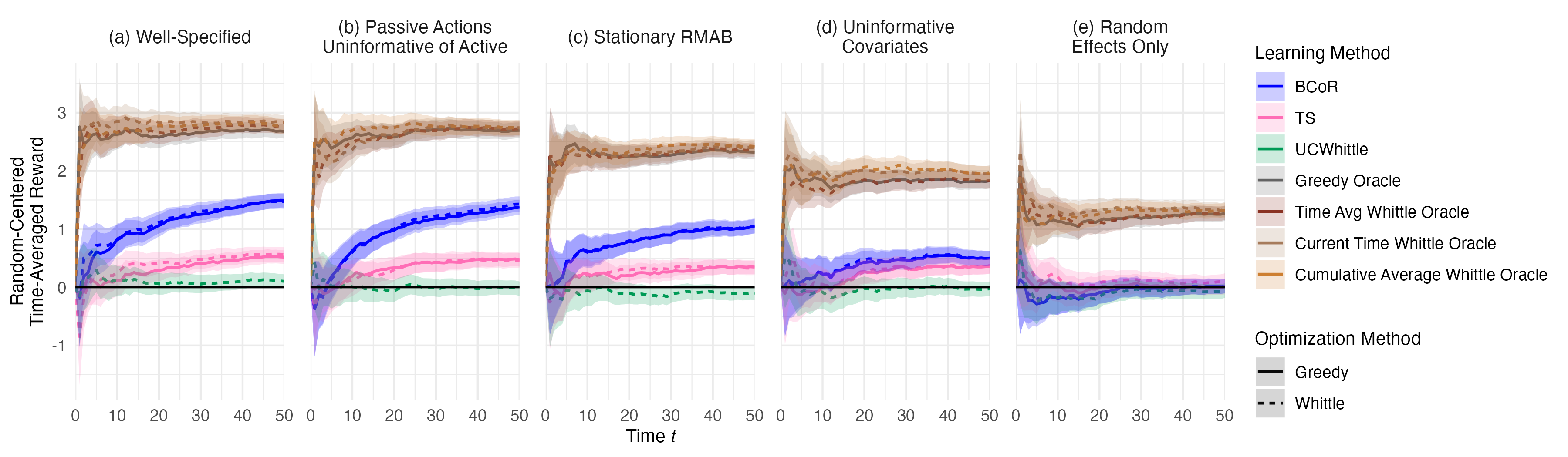}
\vspace{-2em}
\caption{We generate RMAB instances using $N=400,T=50$, and$~B=10$, i.e.,$~B$ is $2.5\%$ of$~N$, across $1{,}000$ random seeds. The covariate matrix $\bm{X}$ was randomly generated with $k=4$ (two continuous covariates and two that are categorical) across the random seeds. \rev{The various RMAB} simulation settings \rev{are detailed in Section~\ref{sec:simulation_exp} and can be summarized as} (a)~a well-specified setting \rev{(no components of  Model~\eqref{final_model} are zero'ed out)}, (b)~a setting where passive actions are uninformative of active actions ($b_0=b_1=0$), (c)~a stationary setting ($\bm{\eta}^{(s,a)} =\rev{\bm{0}, \forall s,a}$), (d)~a setting with uninformative covariate information ($\bm{\mu}_{\bm{\beta}}=0$, all $\bm{\beta}^{(s,a)} =0, \rev{\forall s,a}$), and (e)~a \rev{highly misspecified} setting \rev{,i.e., one where the RMAB instances are stationary with no information sharing between or within the arms}. Lines represent the time-averaged reward of each method \rev{averaged over $1{,}000$ independent instances} with the Random baseline subtracted out. Error bars depict $\pm 2$ SEs. 
}
  \label{sim_exp_all}
\end{figure*}

See Figure~\ref{sim_exp_all} for a version of Figure~\ref{sim_exp} with all Oracles and Greedy policy versions of TS and BCoR present. We find that all three oracles performed comparably in all simulation settings. Additionally, the Greedy versions of BCoR and TS perform similarly to their Whittle counterparts across all experimental settings. This is sensible, since the Greedy oracle often performs comparably to the Whittle oracles as well. 

\clearpage

\subsubsection{Additional Simulation Results}\label{sec:appendix_sim_results}
We also varied the budget and number of covariates to assess performance. See results below.

\begin{figure*}[!hbt]
  \centering
\includegraphics[width=\linewidth]{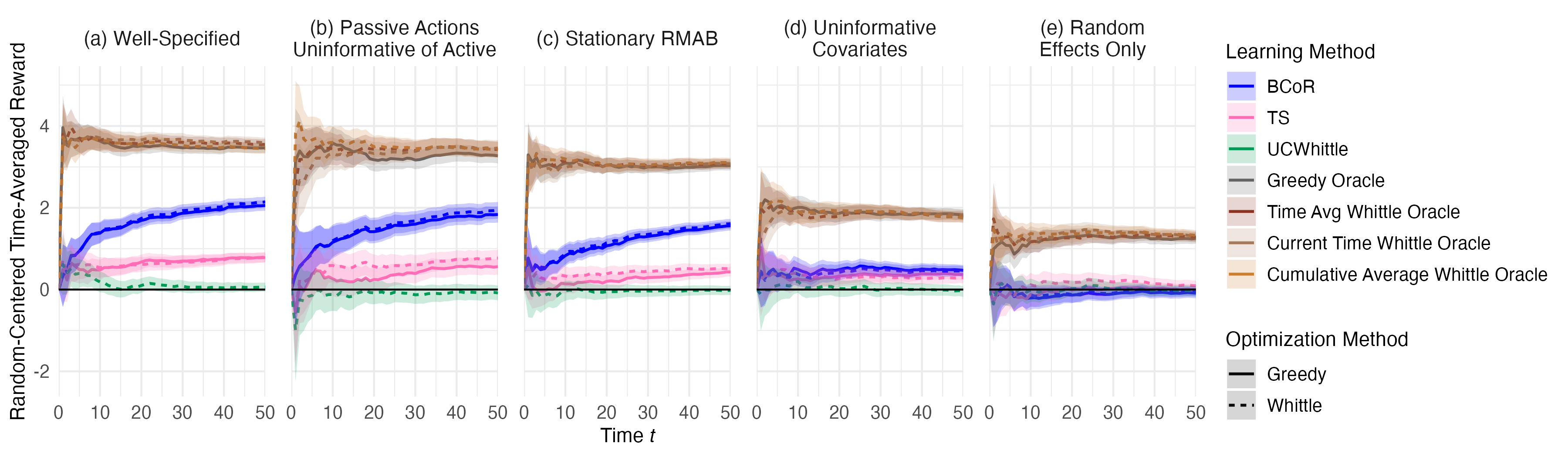}
\vspace{-2em}
\caption{
\textbf{Changing the number of covariates:} We generate RMAB instances using $N=400,T=50$, and$~B=10$, i.e.,$~B$ is $2.5\%$ of$~N$, across $1{,}000$ random seeds. The covariate matrix $\bm{X}$ is randomly generated with $k=8$ (five continuous covariates and three categorical generated as described in Section~\ref{sec:appendix_sim_only}, adding another Bern$(0.5)$ covariate and three additional $\mathcal{N}(0, 1)$ distributed continuous covariates) across the random seeds. \rev{The various RMAB} simulation settings \rev{are detailed in Section~\ref{sec:simulation_exp} and can be summarized as} (a)~a well-specified setting \rev{(no components of  Model~\eqref{final_model} are zero'ed out)}, (b)~a setting where passive actions are uninformative of active actions ($b_0=b_1=0$), (c)~a stationary setting ($\bm{\eta}^{(s,a)} =\rev{\bm{0}, \forall s,a}$), (d)~a setting with uninformative covariate information ($\bm{\mu}_{\bm{\beta}}=0$, all $\bm{\beta}^{(s,a)} =0, \rev{\forall s,a}$), and (e)~a \rev{highly misspecified} setting \rev{,i.e., one where the RMAB instances are stationary with no information sharing between or within the arms}. Lines represent the time-averaged reward of each method \rev{averaged over the $1{,}000$ random seeds} with the Random baseline subtracted out. Error bars depict $\pm 2$ SEs. 
}
  \label{sim_exp_all_8}
\end{figure*}

\begin{figure*}[!hbt]
  \centering
\includegraphics[width=\linewidth]{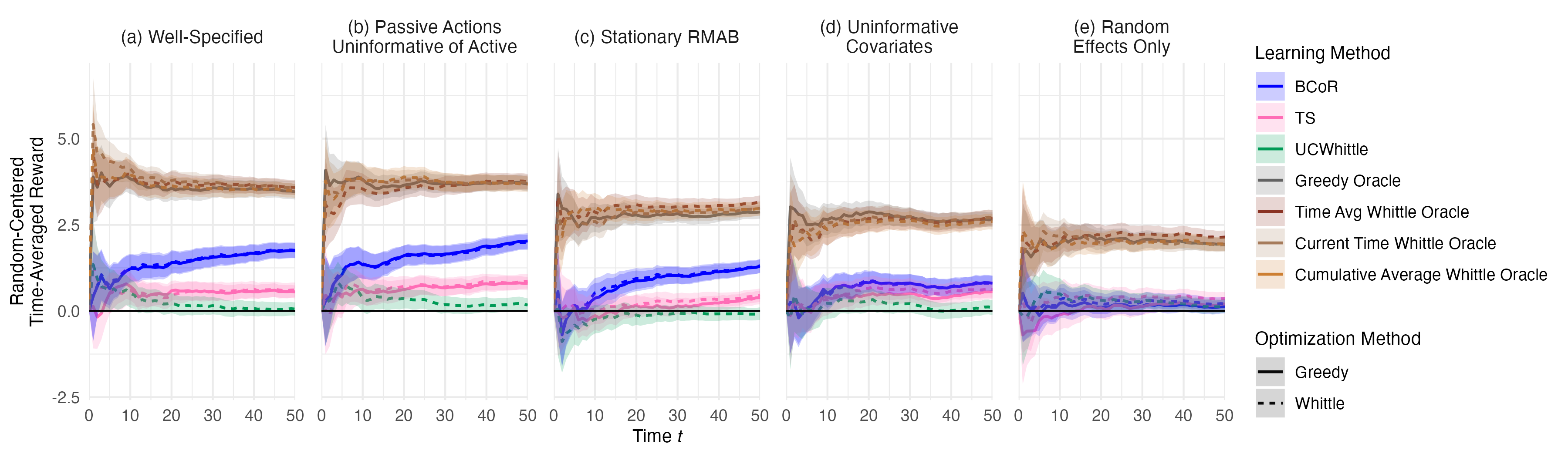}
\vspace{-2em}
\caption{
\textbf{Increasing the budget:} We generate RMAB instances using $N=400,T=50$, and$~B=15$, i.e.,$~B$ is $3.75\%$ of$~N$, across $1{,}000$ random seeds. The covariate matrix $\bm{X}$ was randomly generated with $k=4$ (two continuous covariates and two that are categorical) across the random seeds. \rev{The various RMAB} simulation settings \rev{are detailed in Section~\ref{sec:simulation_exp} and can be summarized as} (a)~a well-specified setting \rev{(no components of  Model~\eqref{final_model} are zero'ed out)}, (b)~a setting where passive actions are uninformative of active actions ($b_0=b_1=0$), (c)~a stationary setting ($\bm{\eta}^{(s,a)} =\rev{\bm{0}, \forall s,a}$), (d)~a setting with uninformative covariate information ($\bm{\mu}_{\bm{\beta}}=0$, all $\bm{\beta}^{(s,a)} =0, \rev{\forall s,a}$), and (e)~a \rev{highly misspecified} setting \rev{,i.e., one where the RMAB instances are stationary with no information sharing between or within the arms}. Lines represent the time-averaged reward of each method \rev{averaged over the $1{,}000$ random seeds} with the Random baseline subtracted out. Error bars depict $\pm 2$ SEs. 
}
  \label{sim_exp_all_b_15}
\end{figure*}

\begin{figure*}[!hbt]
  \centering
\includegraphics[width=\linewidth]{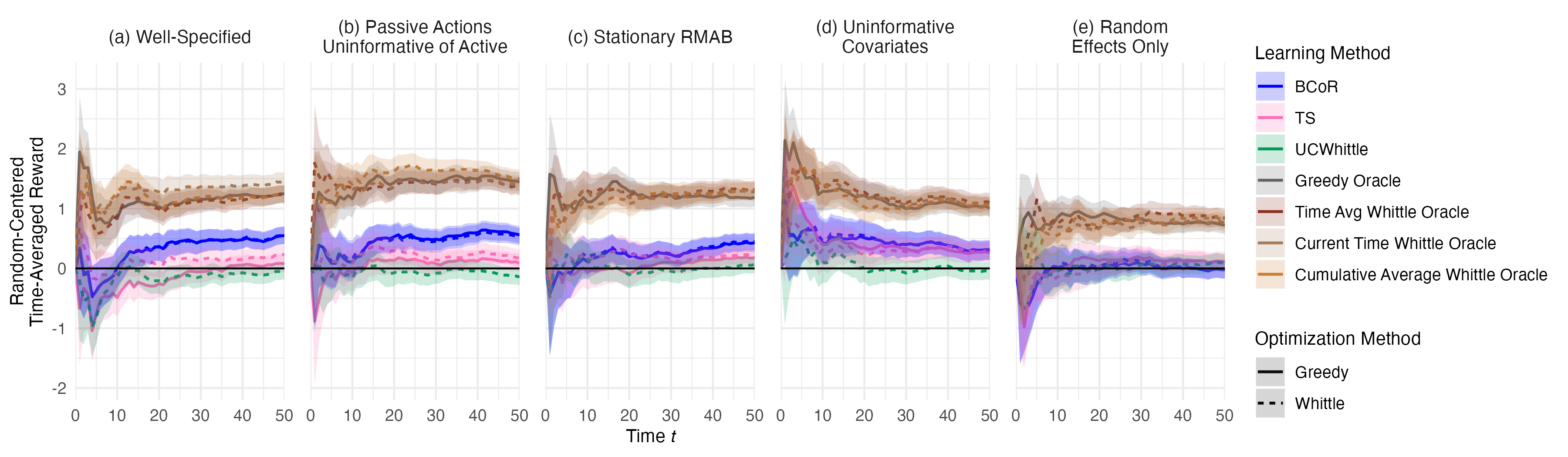}
\vspace{-2em}
\caption{
\textbf{Decreasing the budget:} We generate RMAB instances using $N=400,T=50$, and$~B=5$, i.e.,$~B$ is $1.25\%$ of$~N$, across $1{,}000$ random seeds. The covariate matrix $\bm{X}$ was randomly generated with $k=4$ (two continuous covariates and two that are categorical) across the random seeds. \rev{The various RMAB} simulation settings \rev{are detailed in Section~\ref{sec:simulation_exp} and can be summarized as} (a)~a well-specified setting \rev{(no components of  Model~\eqref{final_model} are zero'ed out)}, (b)~a setting where passive actions are uninformative of active actions ($b_0=b_1=0$), (c)~a stationary setting ($\bm{\eta}^{(s,a)} =\rev{\bm{0}, \forall s,a}$), (d)~a setting with uninformative covariate information ($\bm{\mu}_{\bm{\beta}}=0$, all $\bm{\beta}^{(s,a)} =0, \rev{\forall s,a}$), and (e)~a \rev{highly misspecified} setting \rev{,i.e., one where the RMAB instances are stationary with no information sharing between or within the arms}. Lines represent the time-averaged reward of each method \rev{averaged over the $1{,}000$ random seeds} with the Random baseline subtracted out. Error bars depict $\pm 2$ SEs. 
}
  \label{sim_exp_all_b_5}
\end{figure*}
\clearpage
We also repeated the same experiment with a smaller $N$ and $T$ and varied the budget$~B$. We see similar trends as in the previous simulations, exhibiting BCoR's robustness to these different experimental settings. See results below.

\begin{figure*}[!hbt]
  \centering
\includegraphics[width=\linewidth]{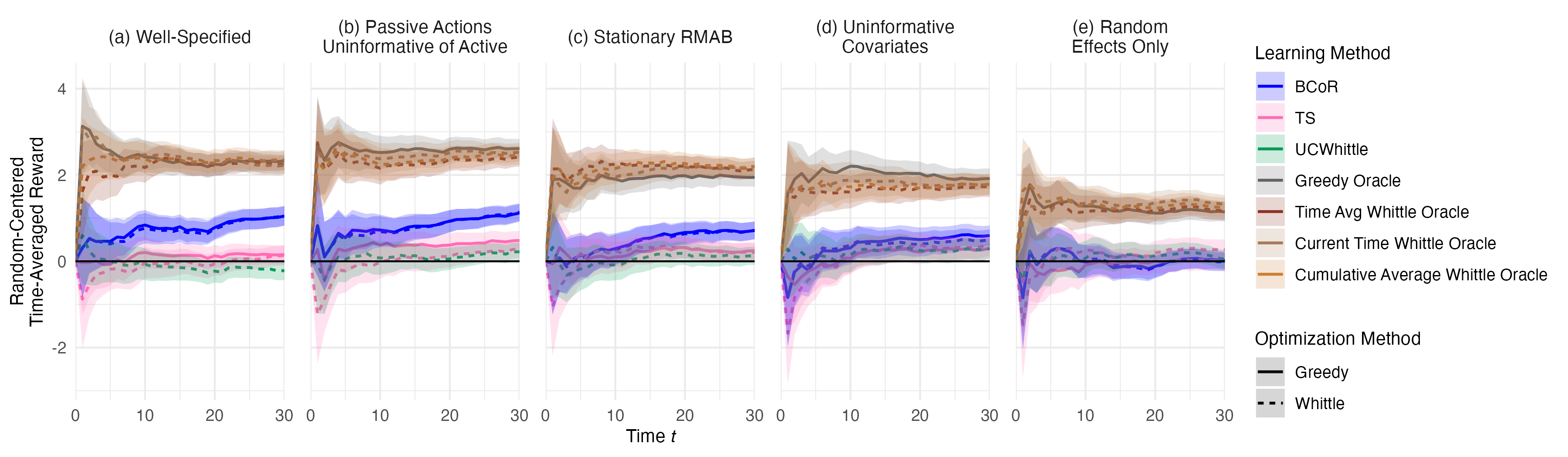}
\vspace{-2em}
\caption{We all generate RMAB instances using $N=300,T=30$, and$~B=10$, i.e.,$~B$ is $3.33\%$ of$~N$. The covariate matrix $\bm{X}$ was randomly generated with $k=4$ (two continuous covariates and two that are categorical) across the random seeds. \rev{The various RMAB} simulation settings \rev{are detailed in Section~\ref{sec:simulation_exp} and can be summarized as} (a)~a well-specified setting \rev{(no components of  Model~\eqref{final_model} are zero'ed out)}, (b)~a setting where passive actions are uninformative of active actions ($b_0=b_1=0$), (c)~a stationary setting ($\bm{\eta}^{(s,a)} =\rev{\bm{0}, \forall s,a}$), (d)~a setting with uninformative covariate information ($\bm{\mu}_{\bm{\beta}}=0$, all $\bm{\beta}^{(s,a)} =0, \rev{\forall s,a}$), and (e)~a \rev{highly misspecified} setting \rev{,i.e., one where the RMAB instances are stationary with no information sharing between or within the arms}. Lines represent the time-averaged reward of each method \rev{averaged over the $400$ random seeds} with the Random baseline subtracted out. Error bars depict $\pm 2$ SEs. 
}
  \label{sim_exp_all_n_300_b_10}
\end{figure*}

\begin{figure*}[!hbt]
  \centering
\includegraphics[width=\linewidth]{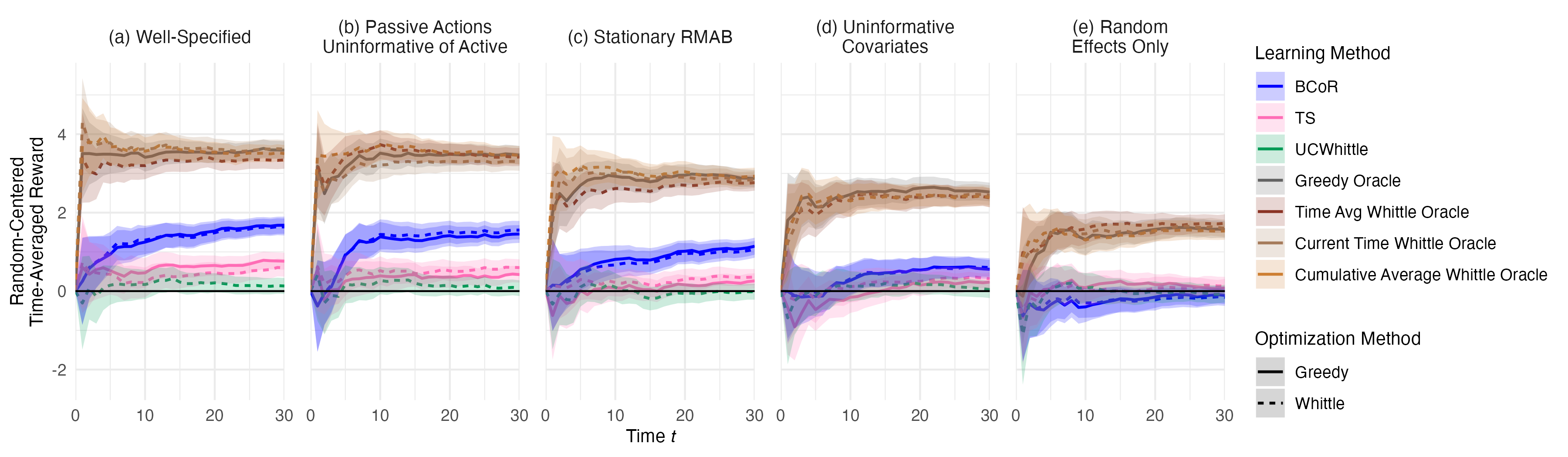}
\vspace{-2em}
\caption{We all generate RMAB instances using $N=300,T=30$, and$~B=15$, i.e.,$~B$ is $5\%$ of$~N$. The covariate matrix $\bm{X}$ was randomly generated with $k=4$ (two continuous covariates and two that are categorical) across the random seeds. \rev{The various RMAB} simulation settings \rev{are detailed in Section~\ref{sec:simulation_exp} and can be summarized as} (a)~a well-specified setting \rev{(no components of  Model~\eqref{final_model} are zero'ed out)}, (b)~a setting where passive actions are uninformative of active actions ($b_0=b_1=0$), (c)~a stationary setting ($\bm{\eta}^{(s,a)} =\rev{\bm{0}, \forall s,a}$), (d)~a setting with uninformative covariate information ($\bm{\mu}_{\bm{\beta}}=0$, all $\bm{\beta}^{(s,a)} =0, \rev{\forall s,a}$), and (e)~a \rev{highly misspecified} setting \rev{,i.e., one where the RMAB instances are stationary with no information sharing between or within the arms}. Lines represent the time-averaged reward of each method \rev{averaged over the $400$ random seeds} with the Random baseline subtracted out. Error bars depict $\pm 2$ SEs. 
}
  \label{sim_exp_all_n_300_b_15}
\end{figure*}

\subsubsection{Transition Dynamic Visualizations for Section~\ref{sec:simulation_exp}}\label{sec:appendix_transition_plots}
Here, we provide visualizations of some of the transition dynamics used in  Section~\ref{sec:simulation_exp}. Note, for the results of Section~\ref{sec:simulation_exp}, we draw a new RMAB instance from the data generating model for each random seed, so the provided plots are examples of a \textit{single} draw from the data generating model; results in Section~\ref{sec:simulation_exp} are averaged over all $1{,}000$ random seeds, hence representing the average reward over $1{,}000$ unique transition dynamics from the data generating model.

\begin{figure*}[h]
  \centering
\includegraphics[width=0.5\linewidth]{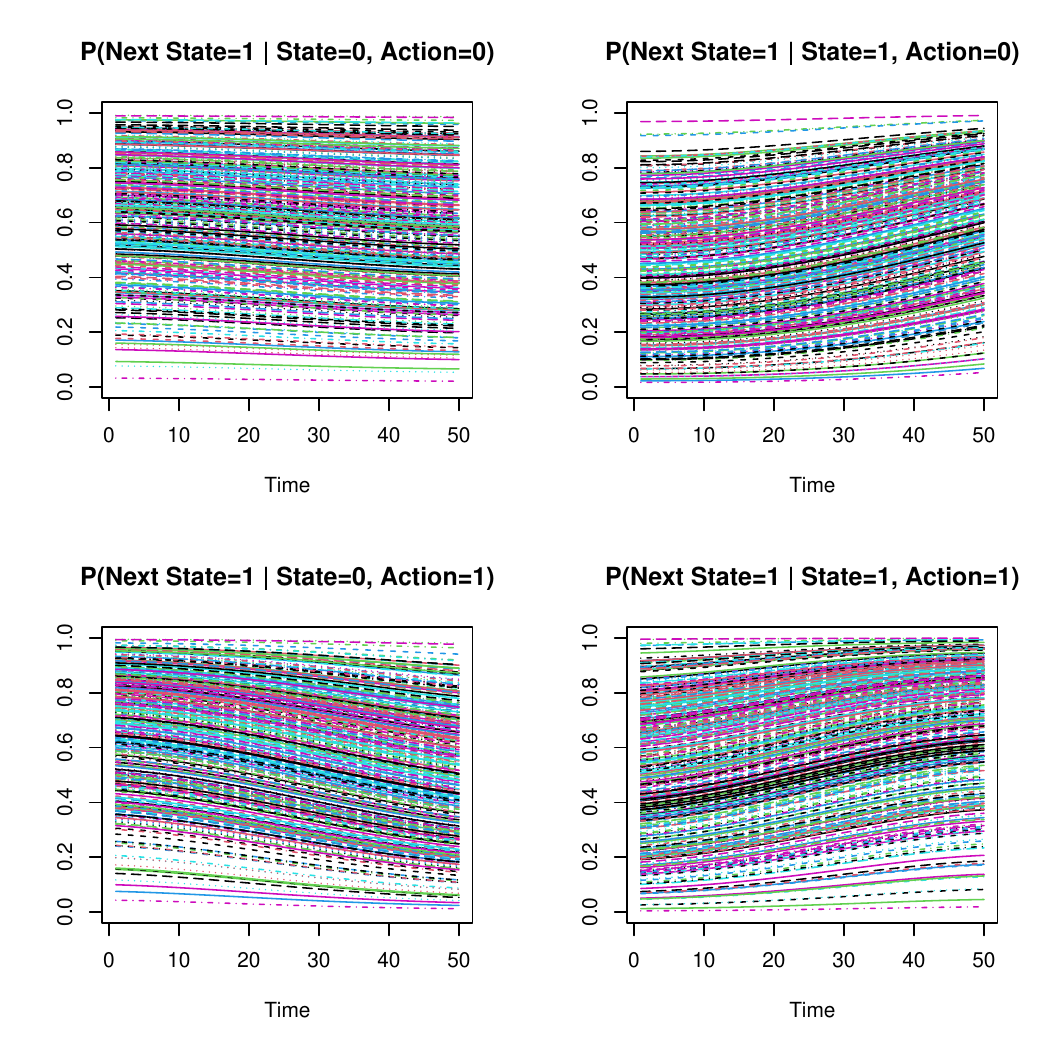}
\caption{
Example of well-specified RMAB instance from Figure~\ref{sim_exp}a.}
\end{figure*}
\begin{figure*}[h]
  \centering
\includegraphics[width=0.5\linewidth]{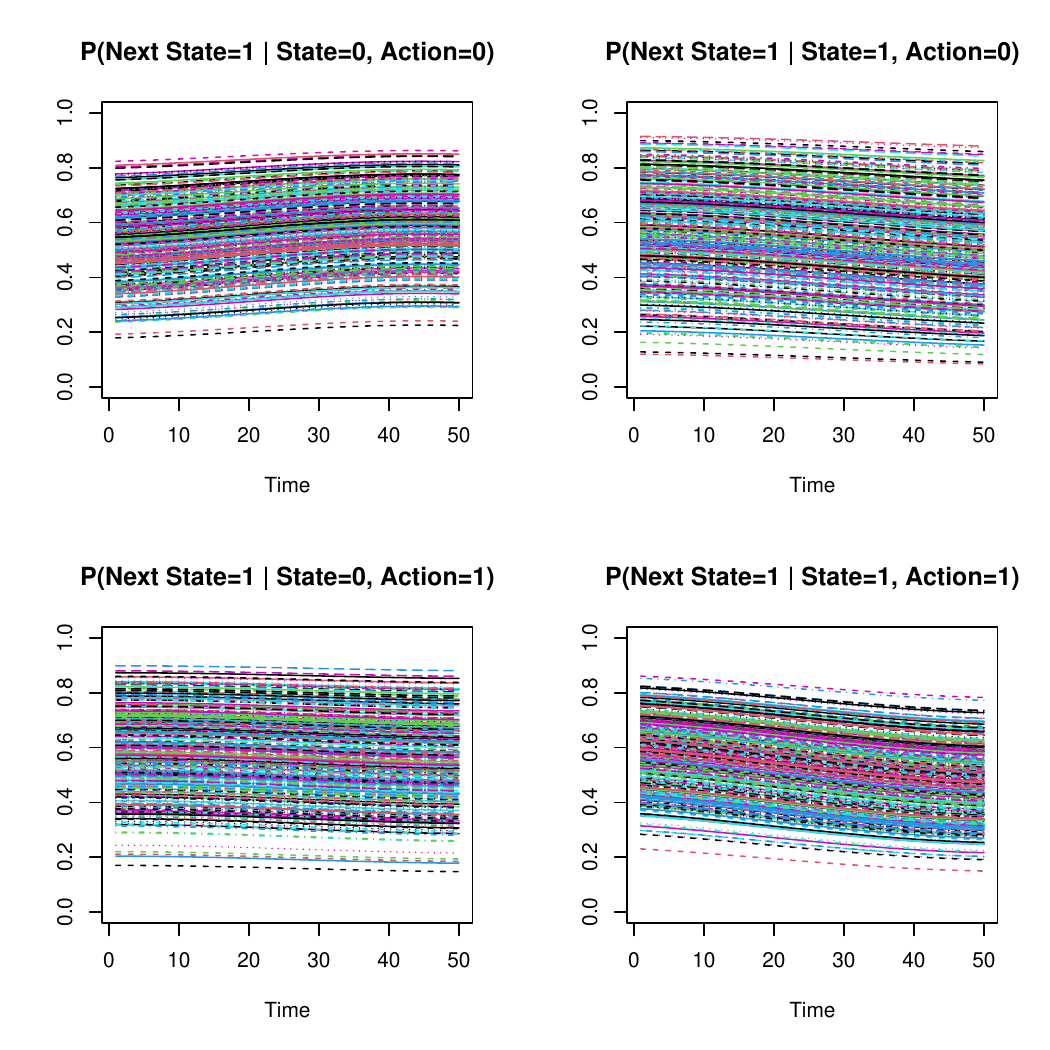}
\caption{
Example of RMAB instance with no information sharing between passive and active actions within an arm from Figure~\ref{sim_exp}b.}
\end{figure*}
\begin{figure*}[h]
  \centering
\includegraphics[width=0.5\linewidth]{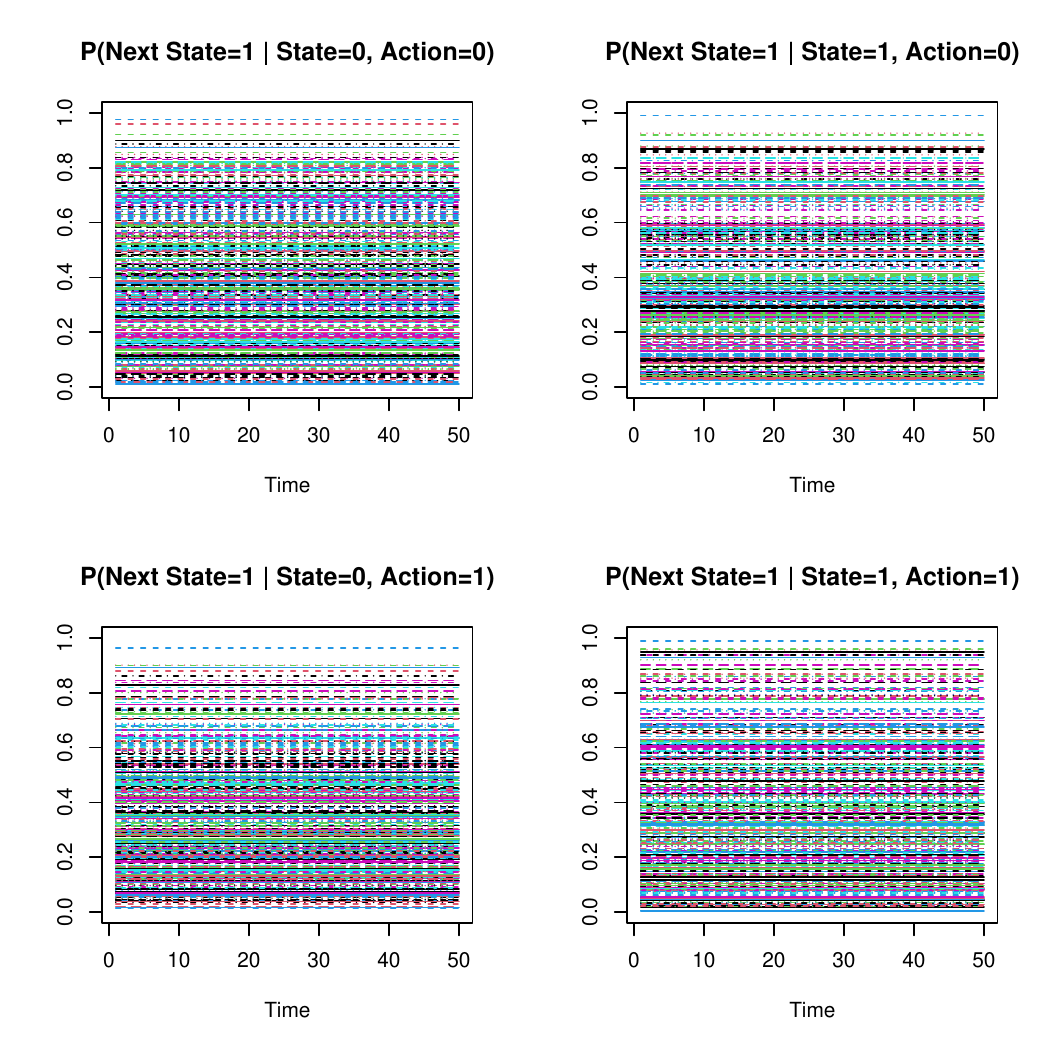}
\caption{
Example of stationary RMAB instance from Figure~\ref{sim_exp}c.}
\end{figure*}

\begin{figure*}[h]
  \centering
\includegraphics[width=0.5\linewidth]{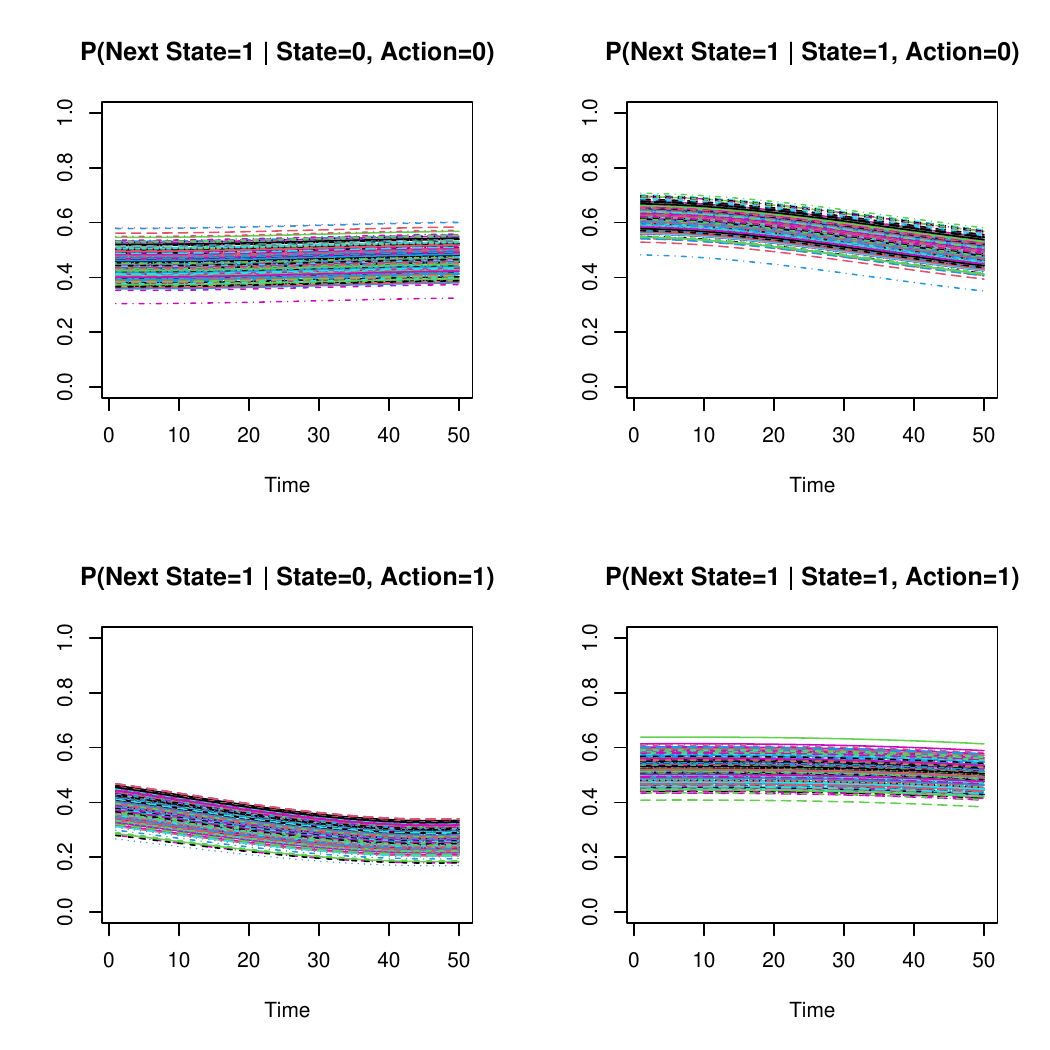}
\caption{
Example of RMAB instance with uninformative covariates from Figure~\ref{sim_exp}d.}
\end{figure*}
\begin{figure*}[h]
  \centering
\includegraphics[width=0.5\linewidth]{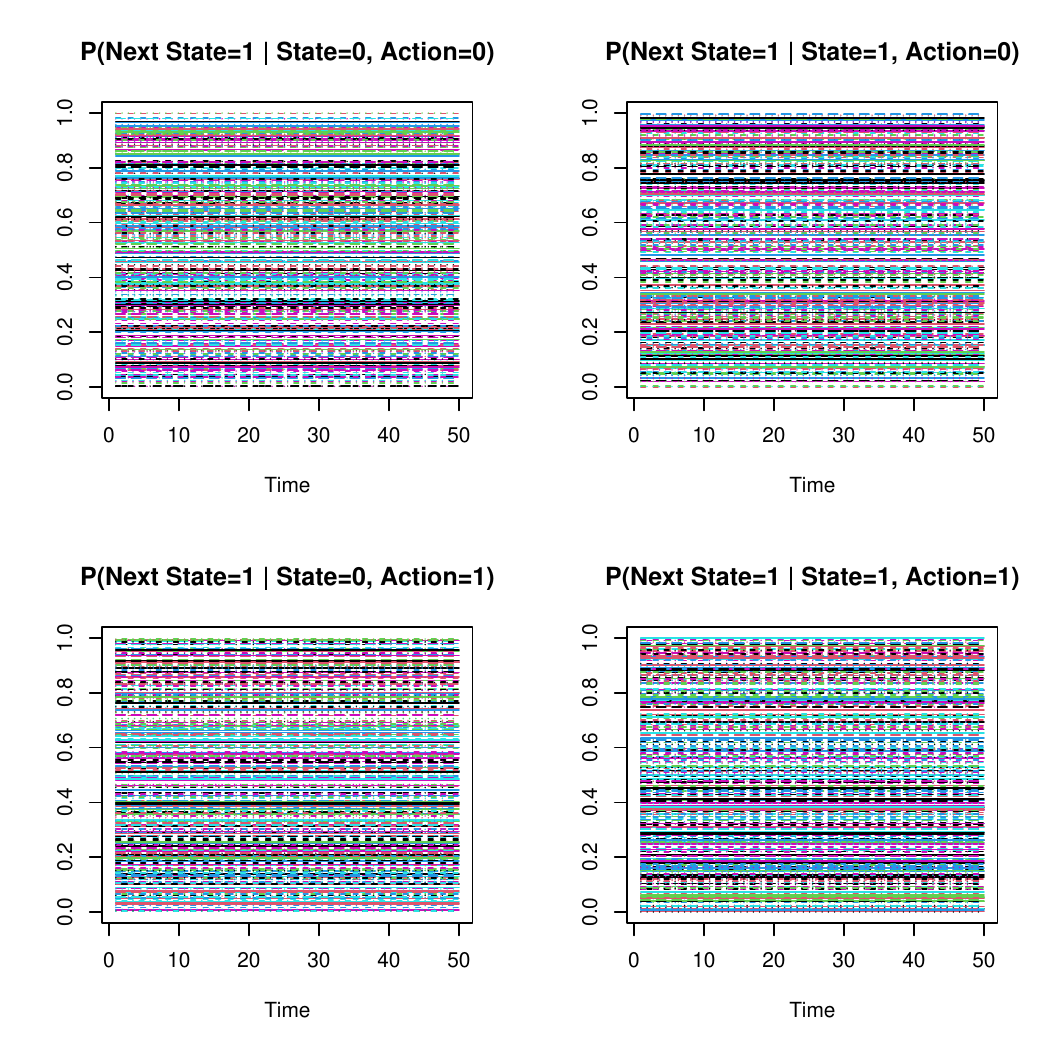}
\caption{
Example of highly misspecified RMAB instance from Figure~\ref{sim_exp}e.}
\end{figure*}

\clearpage

\subsection{Further Details on Section~\ref{sec:real_data_exp}}\label{sec:appendix_armman}
\revtwo{

\subsubsection{Responsible Data Usage}\label{sec:appendix_armman_data_usage}
We provide additional details about data collection, usage and sharing. 
All assets used in the ARMMAN real data example of Section~\ref{sec:real_data_exp} and Appendix~\ref{sec:armman_imp_appendix}, such as the covariate information, the risk scores, and the distributional estimates generated from the risk scores, are owned by ARMMAN and only ARMMAN is allowed to share this information. Beneficiaries for which their data was used to develop the data-driven simulator granted consent to have their anonymized data collected and used for research purposes, with the understanding that such data would not be released publicly. The data collection process, anonymization procedures, and potential use cases were carefully explained to the beneficiaries prior to soliciting their consent and collecting their data. All personally identifiable information about the beneficiaries was removed before sharing with the authors of this paper. 

We, the authors, complied with all ARMMAN data privacy and ethics protocols in the use of the anonymized beneficiary data, such as having read-access only to the anonymized data, restricted usage of the data for only our stated purpose (the analyses of Section~\ref{sec:real_data_exp} and Appendix~\ref{sec:appendix_armman}), and approval from ARMMAN's ethics review committee. In accordance with ARMMAN's data privacy policy, the anonymized covariate information, risk scores, and all scripts used to generate the real data simulator cannot be made publicly available. However, the code, data, and instructions needed to reproduce the fully simulated results \rev{present in Section~\ref{sec:simulation_exp} and Appendix~\ref{sec:appendix_sim_only}} are available via our Github link, including all implementations of the methods under comparison which were used for our ARMMAN data-driven example.
 }

\subsubsection{Data Description and Implementation Details}\label{sec:armman_imp_appendix}
ARMMAN provided anonymized covariate information from $24,011$ beneficiaries enrolled in their maternal health program in 2022. The covariate information included 9 metrics in total, which we can categorize as follows:
\begin{itemize}
  \item Demographic Information:
    \begin{itemize}
        \item `age` (continuous)
        \item `income`, `education`, `phone ownership`, `language` (categorical)
    \end{itemize}
  \item Program Information: 
  \begin{itemize}
      \item `gestational age` (continuous)
      \item `call slot`, `enroll delivery`, `enrollment channel` (categorical)
  \end{itemize}
\end{itemize}

 Through previous analyses, ARMMAN has identified 3 factors that they use to define risk: education, income, and phone ownership. For instance, beneficaries who are illiterate and who do not own their own phone are less likely to engage on average. If a beneficiary falls into the following buckets for education, income, and phone ownership, their risk score increases by $1$:
 \begin{itemize}
    \item Education - ‘Illiterate’, ‘1-5’, ‘6-9’ 
    \item Income - ‘0-5k’, ‘5k-10k’
    \item Phone Owner - ‘family phone’, ‘husband’.
 \end{itemize}   
 Their final risk score is a cumulative count of the number of at-risk metrics they have, so risk scores vary from $0-3$. For instance, a beneficiary who is illiterate and does not own her own phone, but has a household income over $10k$ will have a risk score of $2$. ARMMAN expects patients with risk scores $2$ or $3$ will benefit the most from a live call, although if the risk score is $3$, the beneficiary may not have the means to act on recommendations even when those are given. 
 
 Through previous analyses on prior program runs, ARMMAN has generated distributional estimates of the transition dynamics for each risk score. 
To create the data\rev{-driven} simulator, we first mapped each beneficiary's covariates to her risk score and used ARMMAN's internal distributional estimations of her transition dynamics based on her risk score. This gives us a baseline stationary estimate of each beneficiary's transition dynamics. We use a spline basis model over $1{,}000$ timesteps to generate the non-stationarity using the \texttt{ps} function of \texttt{dlnm}, where the degrees of freedom were set to three and the knots were automatically selected as described and recommended in the \texttt{ps} function documentation of \texttt{dlnm} \cite{dlnm}. To emulate our application area, we run all methods under comparison for only the first $T=40$ timesteps. Hence, the knots used in the spline model to generate the ARMMAN data-driven RMAB instance was set over $1{,}000$ timesteps, but the time horizon provided to BCoR was only $40$. Hence, the resulting RMAB instance has different implied knots and degrees of freedom than the spline model provided to BCoR, meaning the time model is still misspecified for our ARMMAN data-driven example. See Figure~\ref{fig:real_data_transitions} for a visualization of the resulting transition dynamics.

Note, our simulator was developed using real ARMMAN covariates and incorporates ARMMAN's internal estimates of the true transition dynamics based on historical data; these are the same internal estimates that ARMMAN uses for their own internal decision-making and data analyses. Hence, we leveraged all the real data we had access to, given ARMMAN's data exchange policy. Additionally, we verified with ARMMAN's academic collaborators that our simulator outputs reasonably reflected the true data distribution \emph{prior} to any assessment of the algorithms on the simulator. Though we performed this internal assessment, we unfortunately are unable to release any diagnostic plots or analyses, as stipulated by our data exchange agreement with ARMMAN's ethics review committee. 

 BCoR was initialized with the same prior as in Model~\eqref{prior}. TS was initialized with a Uniform prior on $[0, 1]$, and we compute Whittle indices using a discount of $\gamma = 0.9$.
 
\begin{figure*}[h]
  \centering
\includegraphics[width=\linewidth]{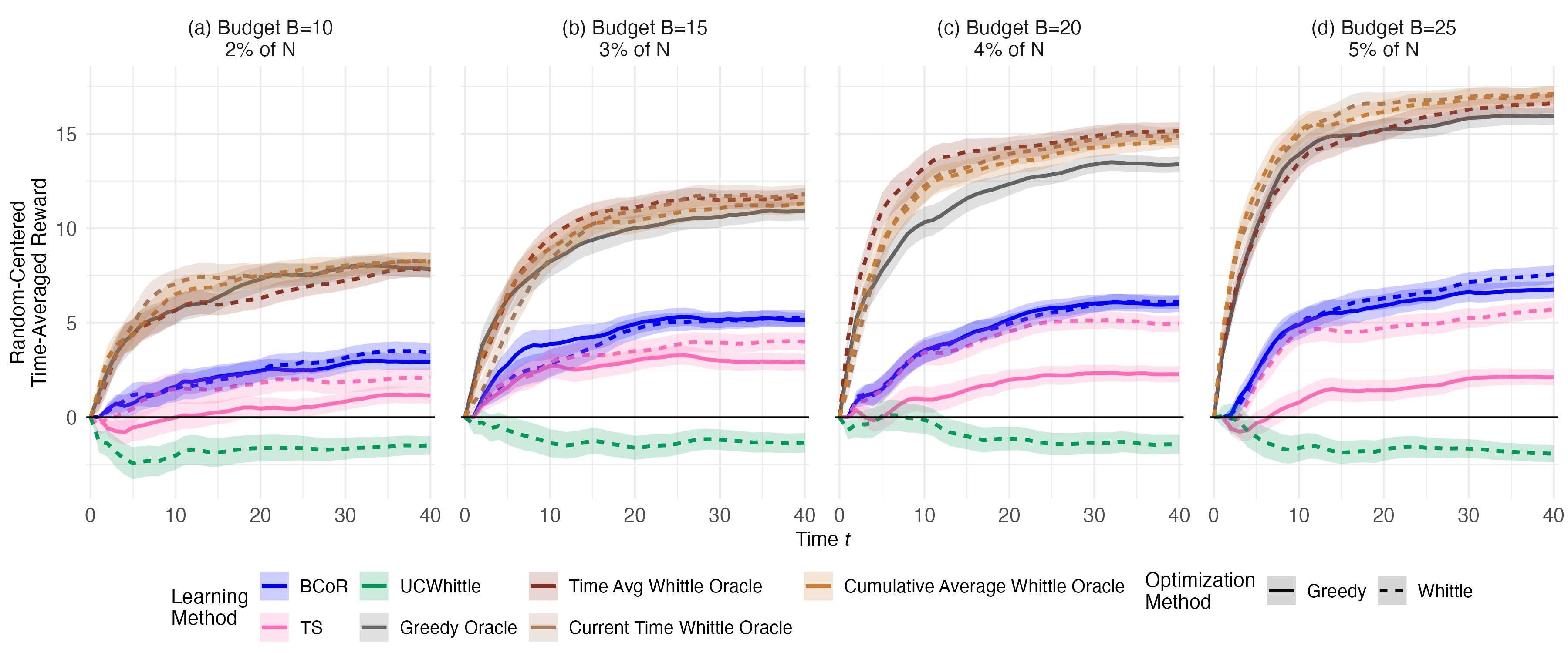}
\caption{
Performance of various methods on the ARMMAN real-data-driven example described in Section~\ref{sec:real_data_exp} with $N=500, T=40$, with varying budget$~B$, where all $B \leq 5\%$ of$~N$ to reflect ARMMAN's true budget constraints. Lines represent the time-averaged reward of each method with the Random baseline subtracted out averaged over $100$ random seeds. Note the top brown and grey methods are oracle approaches with access to the true transitions. Error bars depict $\pm 2$ SEs. 
}
  \label{armman_exp_all}
\end{figure*}

See Figure~\ref{armman_exp_all} for a version of Figure~\ref{armman_exp} with all Oracles and Greedy policy versions of TS and BCoR present. We find that all three oracles performed comparably. Additionally, the Greedy versions of BCoR and TS perform similarly to their Whittle counterparts. This is sensible, since the Greedy oracle performed comparably to the Whittle oracles.

\begin{figure*}[h]
  \centering
\includegraphics[width=\linewidth]{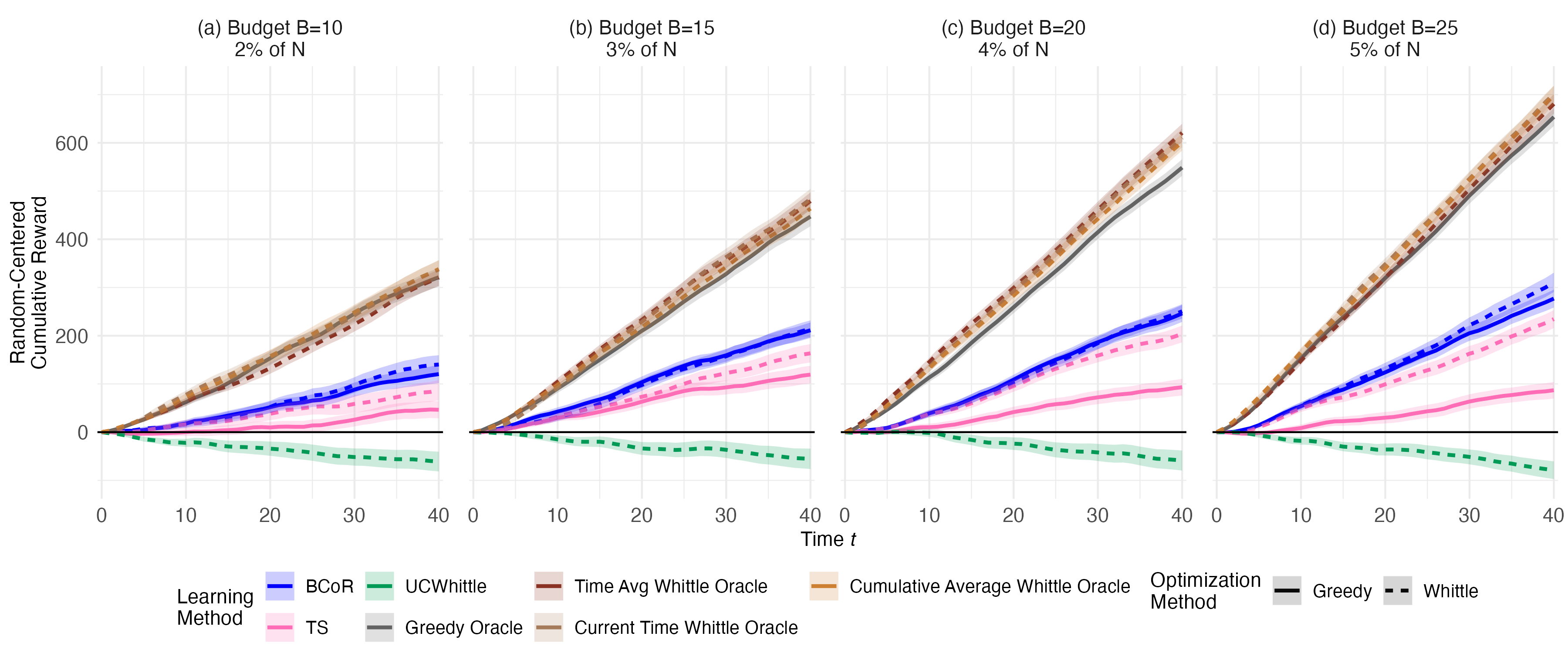}
\caption{
Performance of various methods on the ARMMAN real-data-driven example described in Section~\ref{sec:real_data_exp} with $N=500, T=40$, with varying budget$~B$, where all $B \leq 5\%$ of$~N$ to reflect ARMMAN's true budget constraints. Lines represent the cumulative averaged reward of each method with the Random baseline subtracted out averaged over $100$ random seeds. Note the top brown and grey methods are oracle approaches with access to the true transitions. BCoR consistently outperforms the other approaches. For instance, in $B=10$ setting, BCoR-Whittle had an average random-centered cumulative reward of $140.25$ by the end of the time horizon while TS-Whittle only had $87$, corresponding to an increase of over $61\%$. Error bars depict $\pm 2$ SEs. 
}
  \label{armman_exp_all_cum_avg}
\end{figure*}

\begin{figure}
    \centering
    \includegraphics[width=0.7\linewidth]{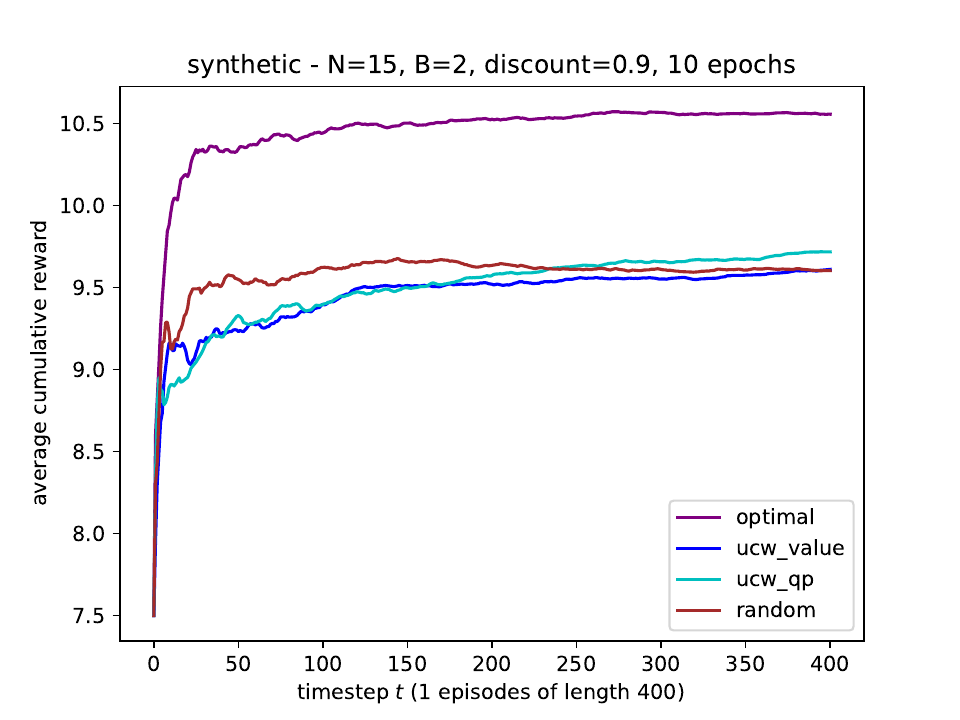}
    \caption{Plot of UCWhittle's performance on a simulated RMAB instance generated using the simulation environment in Section 7 of \cite{ucw}, as implemented in the provided Github repository from \cite{ucw}. ``ucw\_value'' is the method we refer to as UCWhittle in our paper and is the approach primarily presented in \cite{ucw}, as they establish asymptotic regret bounds for this approach. ``ucw\_qp'' is a heuristic version of ucw\_value from \cite{ucw}, which tends to perform comparably to ucw\_value in the short-term across the simulation settings in \cite{ucw} (Given ucw\_value's theoretical guarantees, we use ucw\_value as the method for comparison in our paper). The ``optimal'' line refers to the Whittle index policy which has access to the true transition dynamics. See \cite{ucw} for further implementation details. Note,  ``ucw\_value'' performs worse than random across the short term time horizons, but matches random after some time.}
    \label{fig:ucwhittle_worse}
\end{figure}

\begin{figure}[h]
    \centering
\includegraphics[width=0.8\textwidth]{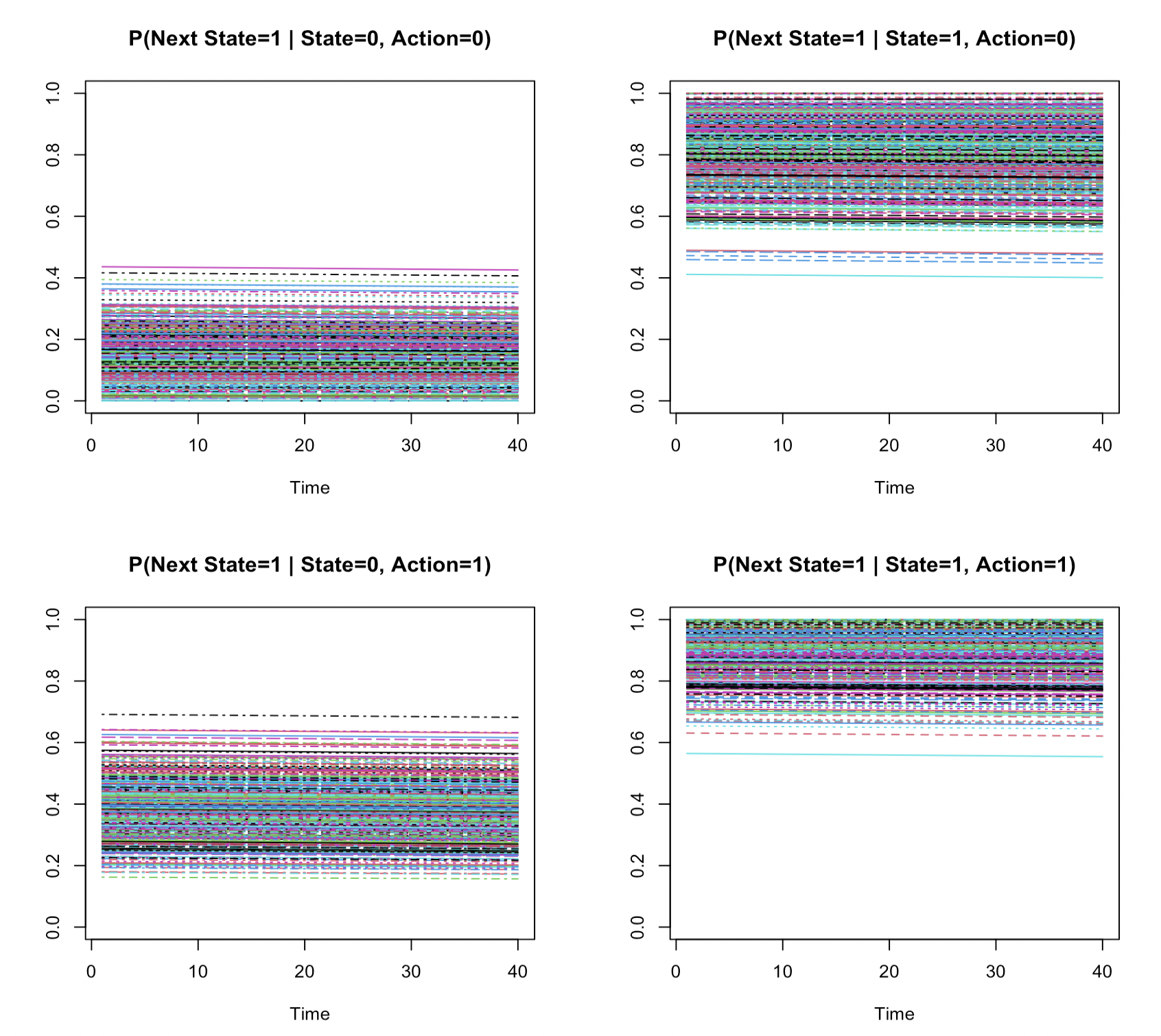}
    \caption{Underlying transition dynamics of the RMAB instance generated from our ARMMAN data-driven simulator, with $N=500$, $T=40$. Each line represents the transition dynamics of an individual arm. Note, the transition dynamics tend to be higher when receiving action$=1$, and that those already in an engaging state $(s=1)$ are more likely to stay in an engaging state than those who are in $s=0$. We confirmed with ARMMAN representatives that these features are reasonable and reflect the dynamics we would expect among actual ARMMAN beneficiaries. Note that there is a only slight amount of non-stationarity which indicates a slight decline in adherence over time.}
    \label{fig:real_data_transitions}
\end{figure}

\clearpage
In our real data experiments, UCWhittle performed worse than random during the entire time horizon of $40$ timesteps. We found that, even when using UCWhittle's own simulation environment from Section 7 of their paper, \cite{ucw}, UCWhittle can perform worse than random in the short term, but eventually matches or outperforms random over time. As UCWhittle's regret bound are asymptotic, it seems that it can have poor finite-sample performance and may require longer time horizons to perform well. See Figure~\ref{fig:ucwhittle_worse} as an example. 
Figure~\ref{fig:ucwhittle_worse} can be reproduced via the repository \texttt{https://github.com/lily-x/online-rmab} and running \texttt{python main.py -N 15 -H 400 -T 1 -B 2 -D synthetic}.

\clearpage


\clearpage
\subsection{Accomodating Continuous Enrollment}\label{sec:appendix_cont_enroll}
In Algorithm~\ref{alg:BCoR}, we choose to set a hyperprior on the random effects. We do so to help our algorithm learn efficiently even in a continuous enrollment setting, where new beneficiaries may join partway through ongoing programs. In principle we do not have any information about the new arms when they first join, but we hope to use information about previously observed arms, particularly those with similar context to the new ones, to better estimate their transition dynamics. Doing so would enable us to quickly and efficiently incorporate them into our intervention allocation. Since the parameters $b_0, b_1$, $\bm{\mu}_{\bm{\beta}}$, the $\bm{\eta}^{(s, a)}$'s, and the $\bm{\beta}^{(s, a)}$'s are shared across all arms, the Bayesian model can immediately use the posterior distribution of the parameters based on previously observed arms and apply them to new arms. However, the $\alpha_j^{(s, a)}$'s are modeled \textit{per arm}. Hence, we cannot directly use the posteriors of the $\alpha_j^{(s, a)}$'s to infer anything about the new arms. 
However, we can interpret the \emph{variances} of the $\alpha_j^{(s,a)}$'s as roughly quantifying the remaining randomness that cannot be explained by the covariate and time effects. We can modify our model to learn these variances so that when a new user enters, our model knows how confident to be in its covariate- and time-based estimate of that user's transition dynamics. We accomplish this by putting a hyperprior on the variance of the random effects $\alpha_j^{(s,a)}$, which we denote by $\tau^2_{\alpha^{(s,a)}}$, effectively treating the variance of the $\alpha_j^{(s,a)}$'s as a parameter in the model.

Specifically, we model the random effects as:
\begin{align*}
\tau^2_{\alpha^{(s, a)}} & \sim \text{Inv-Gamma}(\tau_0, \sigma_0)\nonumber\\
 \alpha_i^{(s, a)} &\sim \mathcal{N}\left(0, \tau^2_{\alpha^{(s, a)}}\right)\nonumber.
\end{align*}
Since $\tau^2_{\alpha_{(s, a)}} $ is shared across all arms for each state-action pair, we can use the posterior distribution of $\tau^2_{\alpha^{(s, a)}}$ given all previously observed data for that state-action pair to infer the distribution of the random effects for the new arms when a new arm is encountered. 
 
  Hence, BCoR can use posterior distributions based on previously observed data to infer the transitions of new (unobserved) beneficiaries that join partway through an ongoing program.  For instance, if a cohort of new beneficiaries joins at some intermediate time point, BCoR has already observed how previous beneficiaries behaved when they were at that point in the program. To provide a concrete example, for ARMMAN's maternal health program, if a cohort of new beneficiaries joins after an initial cohort has already passed their first trimester, BCoR will have already learned from the first cohort how the new beneficiaries are likely to behave in their first trimester. Existing online RL approaches for RMABs cannot use information about previously observed arms to infer the transition dynamics of new arms, so they must incrementally incorporate new arms, about which they initially know nothing, into their existing intervention allocation policy, making them particularly ill-suited for such a continuous enrollment setting. Hence, a continuous enrollment setting would highlight a unique advantage of BCoR, and is often realistic given our motivating application.

\end{document}